\documentclass[11pt]{article}

\usepackage[letterpaper,top=0.72in,bottom=0.72in,left=0.90in,right=0.90in]{geometry}
\usepackage{newtxtext,newtxmath}
\usepackage[T1]{fontenc}
\usepackage{microtype}
\usepackage{graphicx}
\usepackage{booktabs}
\usepackage{array}
\usepackage{multirow}
\usepackage{mathtools,bm}
\usepackage{xcolor}
\usepackage{enumitem}
\usepackage{caption}
\usepackage{subcaption}
\usepackage{natbib}
\usepackage{hyperref}
\usepackage[capitalise,noabbrev]{cleveref}
\usepackage{titlesec}
\usepackage{url}
\usepackage{xspace}
\usepackage{placeins}
\usepackage{float}

\definecolor{pmttblue}{HTML}{0B4778}
\definecolor{pmttteal}{HTML}{007E87}
\definecolor{pmttorange}{HTML}{E58A17}
\definecolor{pmttgray}{HTML}{5A6872}

\hypersetup{
  colorlinks=true,
  linkcolor=pmttblue,
  citecolor=pmttblue,
  urlcolor=pmttteal,
  bookmarksopen=true,
  bookmarksnumbered=true,
  pdfdisplaydoctitle=true,
  pdfauthor={Zixiong Wang and Haiping Ma},
  pdftitle={A Patch is More Than a Token}
}

\titlespacing*{\section}{0pt}{17pt plus 3pt minus 2pt}{7pt plus 1pt minus 1pt}
\titlespacing*{\subsection}{0pt}{12pt plus 2pt minus 1pt}{4pt plus 1pt minus 1pt}
\titlespacing*{\subsubsection}{0pt}{9pt plus 2pt minus 1pt}{3pt}
\titleformat{\section}{\Large\bfseries\color{pmttblue}}{\thesection.}{0.50em}{}
\titleformat{\subsection}{\large\bfseries}{\thesubsection}{0.50em}{}
\titleformat{\subsubsection}{\normalsize\itshape}{\thesubsubsection}{0.50em}{}
\setlist[itemize]{leftmargin=1.45em,itemsep=3pt,topsep=5pt,parsep=0pt}
\setlist[enumerate]{leftmargin=1.55em,itemsep=3pt,topsep=5pt,parsep=0pt}

\newlength{\PMTTAssetWidth}
\newlength{\PMTTFigureFiveHeight}
\newlength{\PMTTFigureBOneHeight}
\newcommand{\PMTTcenterasset}[1]{\makebox[\linewidth][c]{#1}}

\newcommand{\PMTT}{\mbox{\textsc{PMTT}}\xspace}
\newcommand{\PLFS}{Patch-Level Local Feature Space\xspace}
\newcommand{\Center}{\mbox{\textsc{Center}}\xspace}
\newcommand{\Observed}{\mbox{\textsc{Observed}}\xspace}
\newcommand{\Manifold}{\mbox{\textsc{Manifold}}\xspace}
\newcommand{\Visible}{\mbox{\textsc{Visible}}\xspace}
\newcommand{\Hidden}{\mbox{\textsc{Hidden}}\xspace}
\newcommand{\figref}[1]{Fig.~\ref{#1}}
\newcommand{\tabref}[1]{Table~\ref{#1}}

\begin{document}

\vspace*{0.16in}
\begin{center}
{\fontsize{22}{26}\selectfont\bfseries A Patch is More Than a Token\par}
\vspace{12pt}
{\fontsize{12.5}{15}\selectfont Zixiong Wang$^{1}$ \qquad Haiping Ma$^{1}$\par}
\vspace{9pt}
{\fontsize{9.5}{11.5}\selectfont For all inquiries regarding this work, please contact:\par
\textbf{Zixiong Wang}, \href{mailto:zixiongwang.ai@outlook.com}{zixiongwang.ai@outlook.com}\par}
\end{center}
\vspace{16pt}
\begin{center}
\begin{minipage}{0.94\textwidth}
\normalsize
\textbf{Abstract.}
Vision Transformers represent each image patch through a canonical token produced under one observation condition. That token is a powerful interface, but it is still one realized state of the patch. We introduce \PMTT, which establishes two foundations: a patch representation object can extend from one canonical state to a canonical-centered Representation State Family, and the identity-preserving variation of that family opens \PLFS. We develop four central parts of this representation program. Patch Information Acquisition and Dense Orbital Views expose lawful same-patch states. Visible-to-Hidden prediction tests whether observed states reveal structure beyond themselves. Intrinsic analysis identifies low-dimensional, energy-concentrated, patch-specific, predictive geometry. Representation Organization separates the value of acquiring more information from the value of organizing that information, with an analytic \Manifold as the central structured representation.

We realize \PMTT on frozen DINOv3 S/B/L/H+ backbones without neural post-training of the representation. Across two independent 4,096-anchor cohorts, \Manifold explains 70.8--79.6\% of Hidden variation, compared with 55.0--72.3\% for directly \Observed states, and wins on more than 95\% of anchors at every scale. Effective rank falls from about 2.5 to 2.0 while dominant energy concentration rises with backbone scale. Across six visual capabilities and four scales, \Observed exceeds the canonical \Center in 21 of 24 conditions, \Manifold exceeds \Center in 23, and \Manifold exceeds \Observed in 16. The strongest current consequence appears in physical correspondence, where MegaDepth satisfies \Center{} $<$ \Observed{} $<$ \Manifold{} at every scale. These results show that pretrained visual backbones contain richer patch information than the standard single-token interface directly exposes. Code and experimental assets are available at \href{https://github.com/zixiongwang-ai/PMTT}{https://github.com/zixiongwang-ai/PMTT}.
\end{minipage}
\end{center}
\vspace{16pt}
\footnotetext[1]{\color{pmttgray!65}\fontsize{7}{8}\selectfont Shaoxing University}

\section{Introduction}
A Vision Transformer begins by assigning each image patch a token \citep{dosovitskiy2021vit}. This interface has become a basic unit of modern visual representation learning, and self-supervised ViTs have made the token unusually rich: patch features support semantic segmentation, correspondence, depth, video propagation, and other dense tasks with frozen or lightly trained readouts \citep{caron2021dino,oquab2023dinov2,simeoni2025dinov3,amir2022dense}. Yet the representation object itself is usually left unchanged. One patch is observed canonically, one patch token is returned, and downstream methods treat that state as the patch representation.

\PMTT asks whether the canonical token should be the endpoint of that representation. The same physical patch can be re-observed under local changes of scale, crop extent, relative position, aspect, and surrounding context while its spatial identity remains fixed. A frozen backbone then produces multiple lawful states for the same patch. These states are not neighboring patch tokens and they are not synthetic latent samples. They are real backbone responses to identity-preserving observations of the same represented subject.

This shift establishes two equally fundamental foundations. First, the patch representation object extends from one canonical state to a canonical-centered \emph{Representation State Family}. Second, the states and their structured variation open \PLFS, a representation space whose ownership remains with one patch. Once this space is recognized, four questions follow naturally. How should its information be acquired? Do observed states reveal structure that predicts unobserved states? What intrinsic laws organize the space? How should the acquired states be represented for downstream use?

We answer these questions with a clean analytic realization. Dense Orbital Views sample identity-preserving observations around the canonical patch. Sixteen \Visible states construct the representation, and sixteen disjoint \Hidden states test predictive geometry. \Observed directly retains the acquired states. \Manifold uses the same information to build a centered, bounded, low-dimensional continuous organization. The backbone remains frozen, and representation formation trains no generator, adapter, fusion Transformer, or representation MLP. This simplicity is an attribution advantage: any new structure or capability is read from the pretrained representation rather than introduced by post-training.

The intrinsic results are unusually consistent. In two independent 4,096-anchor cohorts and across DINOv3 S, B, L, and H+, \Manifold predicts \Hidden variation substantially better than \Observed. Its pooled coverage rises from about 71\% to 80\%, the anchor-level win rate remains above 95\%, effective rank decreases, and energy concentrates into fewer directions. Random orientation, geometry from another patch, and shared geometry fail to reproduce the effect. The local geometry is therefore not a generic low-rank bias. It belongs to the patch and predicts states that did not participate in its construction.

The downstream evidence establishes that the additional information is usable. Across 24 task-scale conditions, \Observed exceeds \Center in 21 and \Manifold exceeds \Center in 23. \Manifold further exceeds \Observed in 16 conditions, separating acquisition value from organization value. Physical correspondence gives the clearest current result: on MegaDepth, every backbone scale follows \Center{} $<$ \Observed{} $<$ \Manifold. Dense semantics provides a complementary finding. On ADE20K, direct state aggregation can lose value, while \Manifold is best at every scale, showing that more information and better organization are distinct scientific questions.

The central implication is broader than any one downstream task. A pretrained backbone can contain a richer patch representation than the standard single-forward, single-token interface exposes. \PMTT turns that hidden representational richness into a scientific object that can be acquired, predicted, analyzed, organized, and eventually learned during pretraining.

Our contributions are:
\begin{itemize}
  \item We establish a canonical-centered Representation State Family as a patch representation object and formulate \PLFS as the structured space of same-patch observation-conditioned states.
  \item We introduce the Patch Information Acquisition Principle, Dense Orbital Views, and Visible-to-Hidden prediction as a principled route from observation design to predictive patch geometry.
  \item We identify intrinsic laws of \PLFS: low effective dimension, concentrated energy, patch specificity, predictive structure, and systematic geometry scaling across backbone size.
  \item We separate Information Acquisition from Representation Organization and introduce a centered bounded \Manifold that is substantially stronger than direct \Observed states for intrinsic prediction.
  \item We provide a frozen, zero-post-training realization across four DINOv3 scales, two independent intrinsic cohorts, and six visual capabilities, exposing broad downstream gains with minimum-sufficient readers.
\end{itemize}

\FloatBarrier
\newgeometry{top=0.52in,bottom=0.52in,left=0.90in,right=0.90in}
\begin{figure}[H]
  \centering
  \PMTTcenterasset{\includegraphics[width=\PMTTAssetWidth]{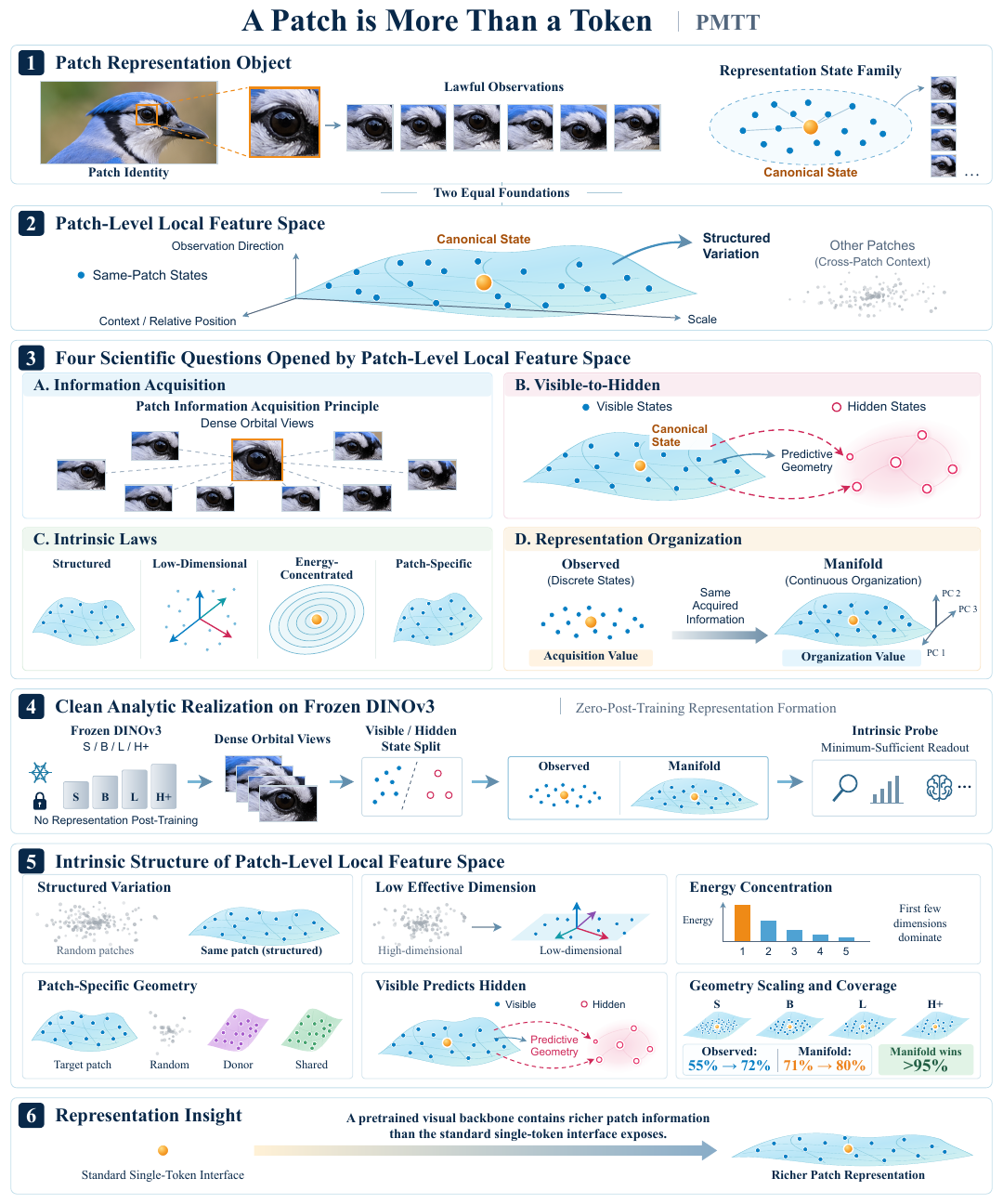}}
  \caption{\textbf{The scientific structure of \PMTT.} A patch retains a fixed spatial identity while lawful local observations produce a canonical-centered Representation State Family. The family opens \PLFS, which motivates four scientific questions: information acquisition, Visible-to-Hidden prediction, intrinsic laws, and representation organization. A clean analytic realization on frozen DINOv3 separates discrete \Observed states from continuous \Manifold organization and reveals low-dimensional, energy-concentrated, patch-specific, predictive geometry. The resulting representation is richer than the standard single-token interface.}
  \label{fig:overview}
\end{figure}
\restoregeometry
\FloatBarrier

\section{PMTT: Patch Representation Object and Patch-Level Local Feature Space}
\subsection{A canonical token is a state, not necessarily the whole object}
Let an image be partitioned into ViT patches, and let $p$ identify one fixed spatial patch. A canonical observation $o_0$ follows the standard patch grid and produces a feature
\begin{equation}
  \bm{c}_p = f(I,o_0,p) \in \mathbb{R}^{d},
\end{equation}
where $f$ is a pretrained visual backbone. The canonical feature is both the conventional patch-token output and the spatial anchor of the representation. \PMTT preserves this role. It changes the endpoint assumption.

An identity-preserving local observation $o$ changes how the patch is seen without changing which physical patch owns the resulting state. The backbone then produces
\begin{equation}
  \bm{s}_p(o)=f(I,o,p).
\end{equation}
The set of lawful states
\begin{equation}
  \mathcal{S}_p=\{\bm{s}_p(o):o\in\mathcal{O}_p\}
\end{equation}
is the Representation State Family of patch $p$. The canonical state remains distinguished, but it no longer exhausts the representation object.

This definition differs from adding neighboring patch tokens. Neighbor tokens describe other spatial identities and can contribute cross-patch context through self-attention. A Representation State Family instead holds the represented subject fixed and varies the observation condition. The ownership of every state remains with patch $p$.

\subsection{Patch-Level Local Feature Space}
The family is more than a collection of points. Its variation defines a local feature space around the canonical state. We call this \PLFS. It includes the canonical center, the realized states produced by lawful observations, and the relations among those states. The scientific object is therefore not only $\bm{c}_p$ and not only a list of alternate features. It is the local organization of same-patch representation states.

This perspective adds a feature-learning axis that is distinct from cross-patch contextualization. Self-attention asks how information from other patches changes the current token. \PMTT asks how the current patch changes when the same identity is observed differently, whether those changes form predictable structure, and whether downstream tasks can use that structure.

The representation depends on three elements: the backbone architecture and weights, the patch identity, and the observation protocol. Standard practice treats the first two as primary and the third as preprocessing. In \PMTT, the observation protocol is part of the representation interface because it determines which lawful states are exposed.

\section{Acquiring and Predicting Patch States}
\subsection{Patch Information Acquisition Principle}
A useful acquisition protocol must expand information without dissolving patch identity. We impose five requirements. Observations remain local to the patch, preserve the underlying image point, cover multiple directions around the canonical view, sample the neighborhood densely enough to expose structure, and avoid task-specific semantic selection. These requirements define the Patch Information Acquisition Principle.

Dense Orbital Views provide a direct realization. Around the canonical patch, we vary crop extent, scale, aspect ratio, surrounding context, and the patch's relative position in the resized crop. The observations are arranged across eight directions with two samples per direction. They are not arbitrary augmentations. Every view is designed to re-observe the same physical patch while changing the local condition under which the frozen backbone represents it.

\subsection{Visible-to-Hidden prediction}
A state family can look structured even if it only interpolates the observations used to construct it. We therefore separate acquisition from evaluation. For each intrinsic anchor, 32 orbital observations are divided into 16 \Visible and 16 \Hidden states. Both sets cover the eight orbital directions, but the \Hidden states never enter representation construction.

The central intrinsic question is whether a representation constructed from \Visible states reduces the error to \Hidden states relative to the canonical center. Let $\mathcal{R}_p$ denote an organized representation and let $\bm{h}_{pk}$ be a \Hidden state. We define
\begin{align}
E_{\mathrm{center}}(p) &= \sum_k \|\bm{h}_{pk}-\bm{u}_p\|_2^2,\\
E_{\mathcal{R}}(p) &= \sum_k \min_{\bm{x}\in\mathcal{R}_p}\|\bm{h}_{pk}-\bm{x}\|_2^2,\\
\mathrm{Coverage}(p) &= 1-\frac{E_{\mathcal{R}}(p)}{E_{\mathrm{center}}(p)},
\end{align}
where $\bm{u}_p=\bm{c}_p/\|\bm{c}_p\|_2$. Positive coverage means that the representation explains unseen state variation beyond the canonical point.

Visible-to-Hidden turns a descriptive cloud into a predictive test. It also defines a natural learning direction: future models can predict withheld orbital states, rather than merely enforcing invariance between views.

\section{Organizing Patch-Level Local Feature Space}
\subsection{Center, Observed, and Manifold}
\Center uses only the canonical state $\bm{u}_p$. \Observed uses the same center and the 16 real \Visible states. It preserves acquired information with the least structural assumption:
\begin{equation}
  \mathcal{R}^{\mathrm{obs}}_p=\{\bm{u}_p,\bm{v}_{p1},\ldots,\bm{v}_{p16}\}.
\end{equation}
For Hidden prediction, the distance is the minimum distance to this discrete set.

\Manifold organizes exactly the same \Visible information. For each normalized \Visible state $\bm{v}_{pj}$, we remove the component parallel to the unit center:
\begin{align}
  \bm{d}_{pj}&=\bm{v}_{pj}-\bm{u}_p,\\
  \bm{r}_{pj}&=\bm{d}_{pj}-\bm{u}_p(\bm{u}_p^{\top}\bm{d}_{pj}).
\end{align}
The residual matrix $R_p=[\bm{r}_{p1},\ldots,\bm{r}_{p16}]$ is summarized by a rank-$r$ factor $A_p\in\mathbb{R}^{d\times r}$, with $r\leq8$, obtained from its dominant singular directions and scales. The local representation is
\begin{equation}
  \mathcal{M}_p=\{\bm{u}_p+A_p\bm{z}:\|\bm{z}\|_2\leq1\}.
  \label{eq:manifold}
\end{equation}
The unit ball makes the representation centered, symmetric, bounded, and continuous. Hidden distance is computed by the constrained least-squares projection onto \cref{eq:manifold}. Basis sign and permutation are immaterial, so the analysis uses basis-invariant spectra, subspace overlap, and distances.

\subsection{Acquisition and organization are different sources of value}
The three representations create a controlled decomposition:
\begin{align}
  \Observed-\Center &:\quad \text{acquisition value},\\
  \Manifold-\Observed &:\quad \text{organization value},\\
  \Manifold-\Center &:\quad \text{total extension}.
\end{align}
The first comparison asks whether additional same-patch states matter. The second holds the acquired information fixed and asks whether explicit geometry adds value. The third measures their combined effect.

The use of a standard analytic factorization is deliberate. A complex learned module could create structure after the fact. Here, the ordinary mathematics is evidence that the structure is already present in the pretrained representation. \Manifold is important not because singular value decomposition is new, but because a patch-specific continuous organization read directly from real states predicts unseen states and improves downstream capability.

\begin{figure}[H]
  \centering
  \PMTTcenterasset{\includegraphics[width=\PMTTAssetWidth]{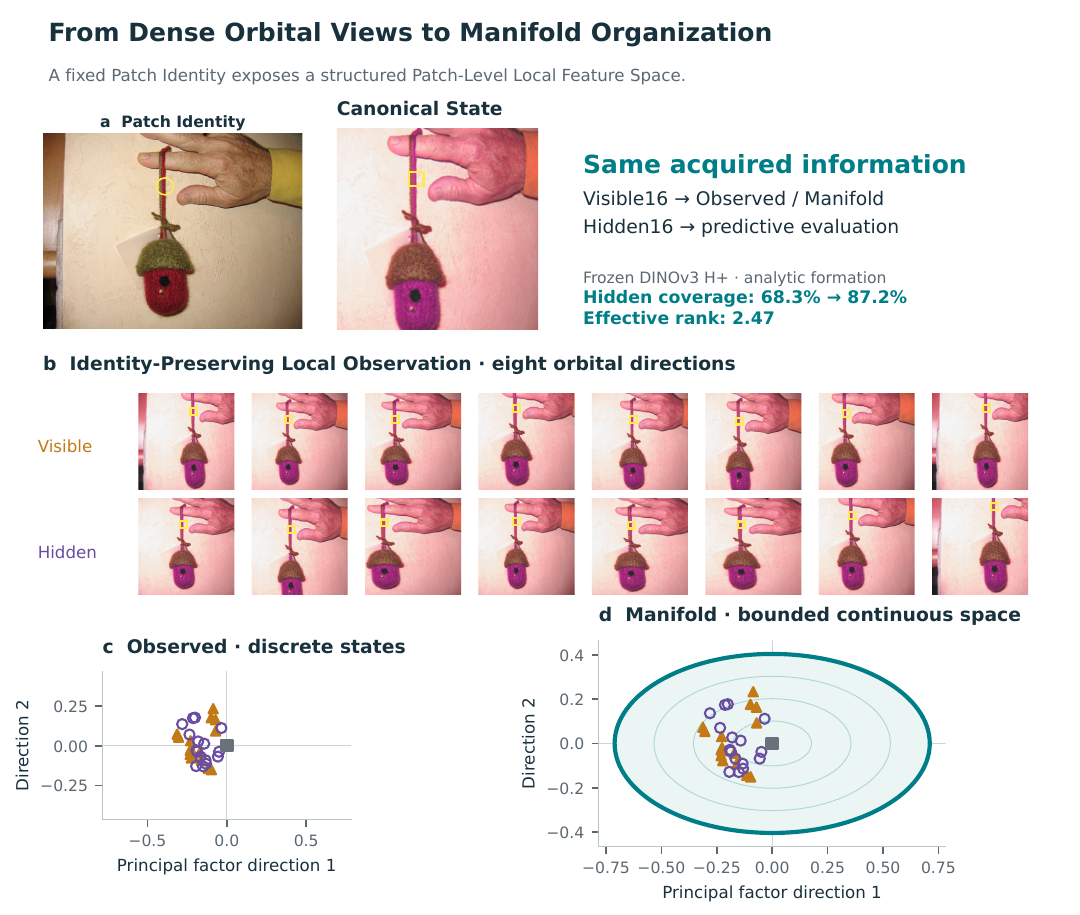}}
  \caption{\textbf{From Dense Orbital Views to Manifold organization.} (a) A frozen DINOv3 example preserves Patch Identity across the original image and Canonical State. (b) Eight orbital directions illustrate Identity-Preserving Local Observation. The full construction uses 16 \Visible states and 16 disjoint \Hidden states. (c) \Observed retains the acquired states as discrete points. (d) \Manifold organizes the same information into a bounded continuous local state space. Triangles denote \Visible states, open circles \Hidden states, and the square the Canonical State.}
  \label{fig:construction}
\end{figure}
\FloatBarrier

\section{Relation to Existing Representation Learning}
\paragraph{Patch tokens and token-set modifications.}
ViT established the patch token as the basic visual input unit \citep{dosovitskiy2021vit}. Later work modifies token sets for different goals. Register tokens provide dedicated non-spatial computational slots and remove artifacts from patch feature maps \citep{darcet2024registers}. TokenLearner adaptively extracts a small set of informative tokens \citep{ryoo2021tokenlearner}, token merging combines similar tokens for efficiency \citep{bolya2023tome}, and mixed-resolution tokenization reallocates spatial resolution across the image \citep{ronen2023mixed}. Recent SubViT selectively subdivides discriminative patches into multiple subtokens and trains a router for fine-grained recognition \citep{zhu2026subvit}. These methods alter the token sequence, spatial granularity, or computational allocation. \PMTT preserves the canonical patch identity and instead expands the representation object of one patch through multiple states generated by lawful re-observation.

\paragraph{Multi-view self-supervision and invariance.}
Multi-view self-supervised learning uses crops and transformations to define training signals. SwAV introduced multi-crop assignment prediction \citep{caron2020swav}, DINO uses global and local crops in self-distillation \citep{caron2021dino}, and iBOT and MAE combine view learning with masked prediction or reconstruction \citep{zhou2022ibot,he2022mae}. EsViT and VICRegL add region-level or location-consistent local objectives \citep{li2022esvit,bardes2022vicregl}, CroCo learns cross-view completion from distinct camera views \citep{weinzaepfel2022croco}, and AugSelf explicitly preserves augmentation parameters \citep{lee2021augself}. DINOv2 and DINOv3 scale self-supervision into general-purpose frozen visual backbones \citep{oquab2023dinov2,simeoni2025dinov3}. These methods use views to shape a learned representation. \PMTT instead treats the post-training differences among lawful observations of one patch as the representation object itself, predicts withheld states, and organizes their geometry rather than collapsing it.

\paragraph{Equivariant and transformation-aware representations.}
Equivariant SSL explicitly preserves transformation information during training. E-SSL predicts applied transformations \citep{dangovski2022equivariant}. SIE separates invariant and equivariant components using a learned predictor \citep{garrido2023sie}. Transformation learning and equivariance-coherent reconstruction further develop learned transformation representations and intermediate transformed states \citep{yu2024stl,wang2026equivariance}. \PMTT shares the motivation that transformation-induced variation can be useful, but its object and evidence differ. It does not require transformation labels, a learned transformation operator, or backbone training. It measures the actual observation-conditioned state geometry of a frozen patch representation and tests whether visible geometry predicts hidden states.

\paragraph{Sub-token detail and repeated inference.}
Stochastic Resonance Transformers apply sub-token input translations, inverse-align feature maps, and aggregate them to recover finer spatial detail without fine-tuning \citep{lao2024srt}. FeatUp similarly uses transformed low-resolution feature views to learn a higher-resolution feature field \citep{fu2024featup}. Their endpoint is a spatially refined or upsampled feature map. \PMTT keeps each same-patch state as a member of a representation family, separates acquisition from organization, and evaluates the resulting local geometry through Visible-to-Hidden prediction. Its endpoint is a richer patch representation object rather than spatial upsampling.

\paragraph{Frozen ViT features for dense matching.}
Self-supervised ViT patch features already serve as strong dense descriptors \citep{amir2022dense}. Geometry-aware correspondence methods expose orientation and symmetry limitations of foundation-model features, while more recent work trains adapters or pseudo-label pipelines to improve matching \citep{zhang2024leftright,dunkel2025diy}. \PMTT asks a complementary question: how much correspondence information is already present in a frozen backbone but hidden by the canonical token interface? Its clean readout makes the representation interface, rather than a trained matcher, the main experimental variable.

\section{A Clean Analytic Realization on Frozen DINOv3}
\subsection{Backbones and observations}
We evaluate DINOv3 ViT-S, ViT-B, ViT-L, and ViT-H+ pretrained on LVD-1689M \citep{simeoni2025dinov3}. Their feature dimensions are 384, 768, 1024, and 1280. All use patch size 16. Backbone parameters remain frozen throughout representation formation and downstream evaluation. Features are produced in inherited BF16 autocast and converted to FP32 for geometry and readers. TF32 is disabled.

For each intrinsic anchor, we retain the canonical observation and 32 Dense Orbital Views. Sixteen \Visible observations construct \Observed and \Manifold. Sixteen \Hidden observations are held out. The Original Cohort and Independent Holdout Cohort each contain 4,096 mutually fixed anchors and remain separate throughout analysis.

\subsection{Minimum-sufficient representation-native readers}
Downstream evaluation uses low-capacity readers designed around each representation rather than one generic high-capacity decoder. The \Center reader accesses canonical feature similarities or a Center-only task axis. The \Observed reader summarizes responses across the 16 real states using fixed symmetric statistics such as the mean, maximum, top-$k$ mean, variation, and calibrated log-mean-exp. The \Manifold reader uses basis-invariant quantities such as bounded distance, subspace distance, explained ratio, coefficient norm, spectrum concentration, or the radius of a downstream task axis within the local factor.

SPair and DAVIS fit small regularized scalar readers on their official train-side data. MegaDepth and ScanNet use the same candidate grids, anchor identities, one-to-one assignment, and local refinement for all representations. NYUv2 and ADE20K first fit a Center-only linear depth or semantic axis at each scale. \Observed and \Manifold then use only a small class-independent scalar mixture of state-family or geometry quantities. The same task axis is shared across the three representations, preventing the downstream model from learning separate high-capacity solutions.

\subsection{Tasks, metrics, and statistical units}
We evaluate six capabilities. SPair-71k measures semantic keypoint correspondence with PCK@0.1 of the object bounding box \citep{min2019spair}. DAVIS-2017 measures first-frame video object propagation with region $J$, contour $F$, and their mean \citep{perazzi2016davis,ponttuset2017davis}. MegaDepth-1500 and ScanNet-1500 measure outdoor and indoor physical correspondence with precision within 10 resized target pixels, and additionally report pose metrics \citep{li2018megadepth,dai2017scannet}. NYUv2 measures dense depth with AbsRel and standard secondary metrics \citep{silberman2012nyu}. ADE20K measures 150-class semantic segmentation with dataset-level mIoU \citep{zhou2017ade20k}.

The primary matrix contains four scales, six tasks, and three representations, giving 72 scores. Paired intervals use the natural evaluation unit of each task and 2,000 bootstrap draws. For MegaDepth and ScanNet, descriptive means use all 1,500 pairs, while paired inference uses 1,372 untouched pairs after excluding the fixed calibration-overlap identities. No cross-task aggregate score is constructed because the metrics have different meanings.

\subsection{Attribution controls}
Intrinsic controls preserve selected low-level properties while breaking the scientific relation of interest. Random geometry preserves rank, spectrum, energy, and bound but rotates the local directions. Donor geometry uses another patch's factor. Shared geometry uses structure estimated without the current patch-specific response. Full-32 uses both Visible and Hidden states to construct the factor and serves only as a diagnostic ceiling.

Downstream DonorObserved replaces same-patch \Visible information with deterministic donor states while keeping the reader fixed. DestroyedManifold preserves the factor spectrum and bound while destroying tangent orientation. Controls support attribution. They do not enter the 72 primary cells.

\section{Intrinsic Laws of Patch-Level Local Feature Space}
\subsection{Manifold predicts unseen states at every scale}
\begin{figure}[H]
  \centering
  \PMTTcenterasset{\includegraphics[width=\PMTTAssetWidth]{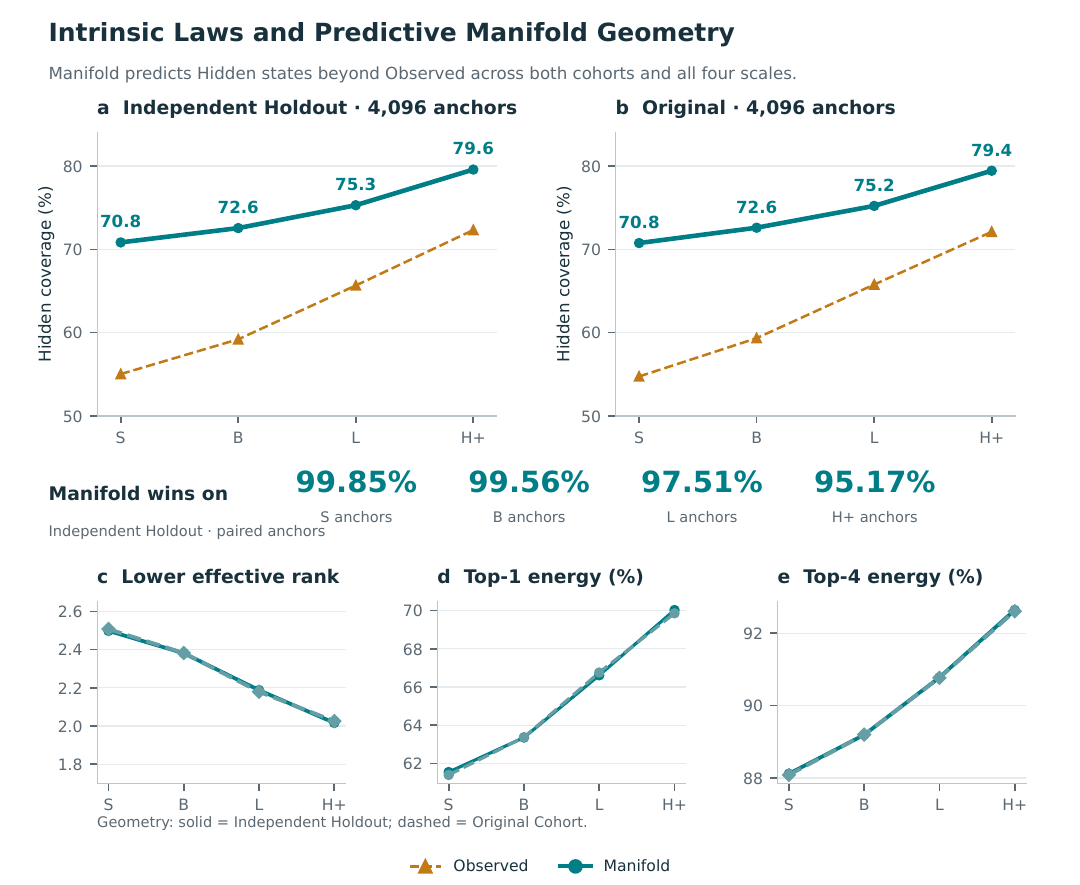}}
  \caption{\textbf{Intrinsic Laws and predictive Manifold geometry.} \Manifold predicts \Hidden states substantially better than \Observed across both independent cohorts and all four frozen DINOv3 scales. Supporting panels show decreasing effective rank and increasing energy concentration. Solid lines denote the Independent Holdout Cohort and dashed lines the Original Cohort.}
  \label{fig:intrinsic}
\end{figure}

\figref{fig:intrinsic} and \tabref{tab:intrinsic} show the central intrinsic result. In the Independent Holdout Cohort, \Observed coverage rises from 55.03\% at S to 72.32\% at H+. \Manifold rises from 70.84\% to 79.58\%. The organization gain is positive at all scales: 15.81, 13.35, 9.64, and 7.26 percentage points. The corresponding anchor-level win rates are 99.85\%, 99.56\%, 97.51\%, and 95.17\%.

The Original Cohort independently reproduces the same ordering and nearly the same absolute values. This agreement is important because the result is not carried by one fixed image set. At every scale and in both cohorts, a geometry constructed only from \Visible states explains \Hidden variation better than the discrete set of observed points.

\begin{table}[H]
  \centering
  \PMTTcenterasset{\includegraphics[width=\PMTTAssetWidth]{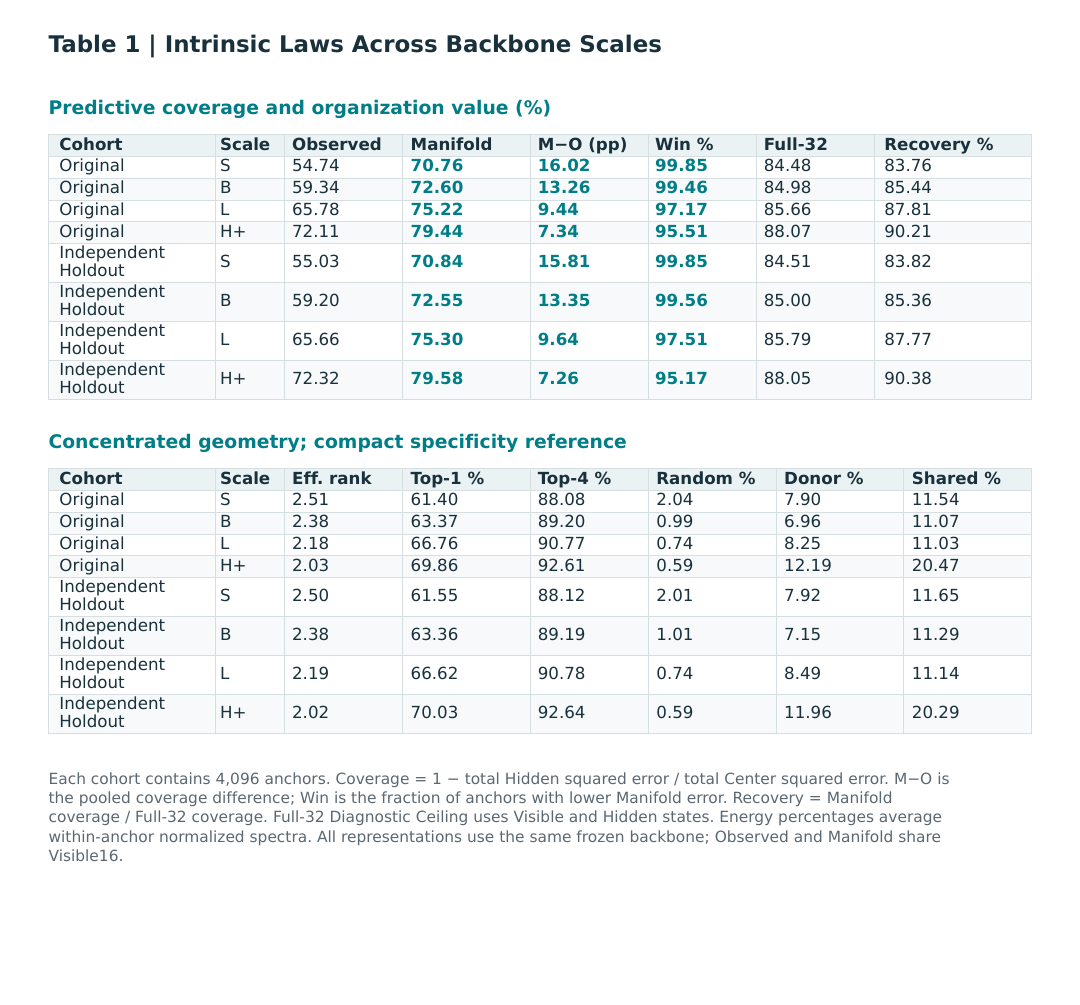}}
  \caption{\textbf{Intrinsic Laws across backbone scales.} Each cohort contains 4,096 anchors. Coverage measures the reduction in total \Hidden squared error relative to the Canonical State. Full-32 uses \Visible and \Hidden states and is a diagnostic ceiling. Energy percentages average within-anchor normalized spectra.}
  \label{tab:intrinsic}
\end{table}
\FloatBarrier

\subsection{The local space is low-dimensional and energy-concentrated}
The same scaling trend appears in the geometry itself. In the Independent Holdout Cohort, effective rank falls from 2.50 at S to 2.02 at H+. The top singular direction increases from 61.55\% to 70.03\% of normalized energy, and the top four directions increase from 88.12\% to 92.64\%. The Original Cohort gives nearly identical values. Although the feature dimension grows from 384 to 1280, the observation-conditioned variation becomes more concentrated rather than more diffuse.

This result is stronger than a statement that rank eight is sufficient. Most of the predictive geometry is effectively carried by about two dominant directions, and the retained rank leaves room for less energetic structure without forcing it into the principal axes.

\subsection{The geometry is patch-specific}
Correct \Manifold coverage is separated from the controls by a large margin. In the Independent Holdout Cohort, Random coverage falls from about 2.0\% at S to 0.6\% at H+. Donor geometry ranges from about 7.9\% to 12.0\%, and Shared geometry from 11.6\% to 20.3\%. These controls can retain low rank or substantial energy, but they do not recover the target patch's \Hidden states.

The result localizes the structure to the patch. A generic low-dimensional subspace, a factor from another patch, or a shared response cannot substitute for the correct directions. The specific orientation of the local geometry carries information.

\subsection{Backbone scale strengthens both acquisition and organization}
\Observed and \Manifold coverage both rise with backbone scale. A stronger backbone therefore exposes more usable same-patch information through the acquired states and organizes that information into a more compact predictive geometry. \Manifold also recovers an increasing fraction of the Full-32 ceiling, from 83.82\% at S to 90.38\% at H+ in the Independent Holdout Cohort. The ceiling remains higher, which leaves acquisition headroom, but the visible-derived representation already captures most of the available rank-8 local structure.

Together, these results establish the intrinsic laws of \PLFS in the evaluated backbones: state variation is structured, low-dimensional, energy-concentrated, patch-specific, predictive, and increasingly compact across scale.

\section{Capability Consequences}
\subsection{Additional same-patch information has broad value}
\begin{figure}[H]
  \centering
  \PMTTcenterasset{\includegraphics[width=\PMTTAssetWidth]{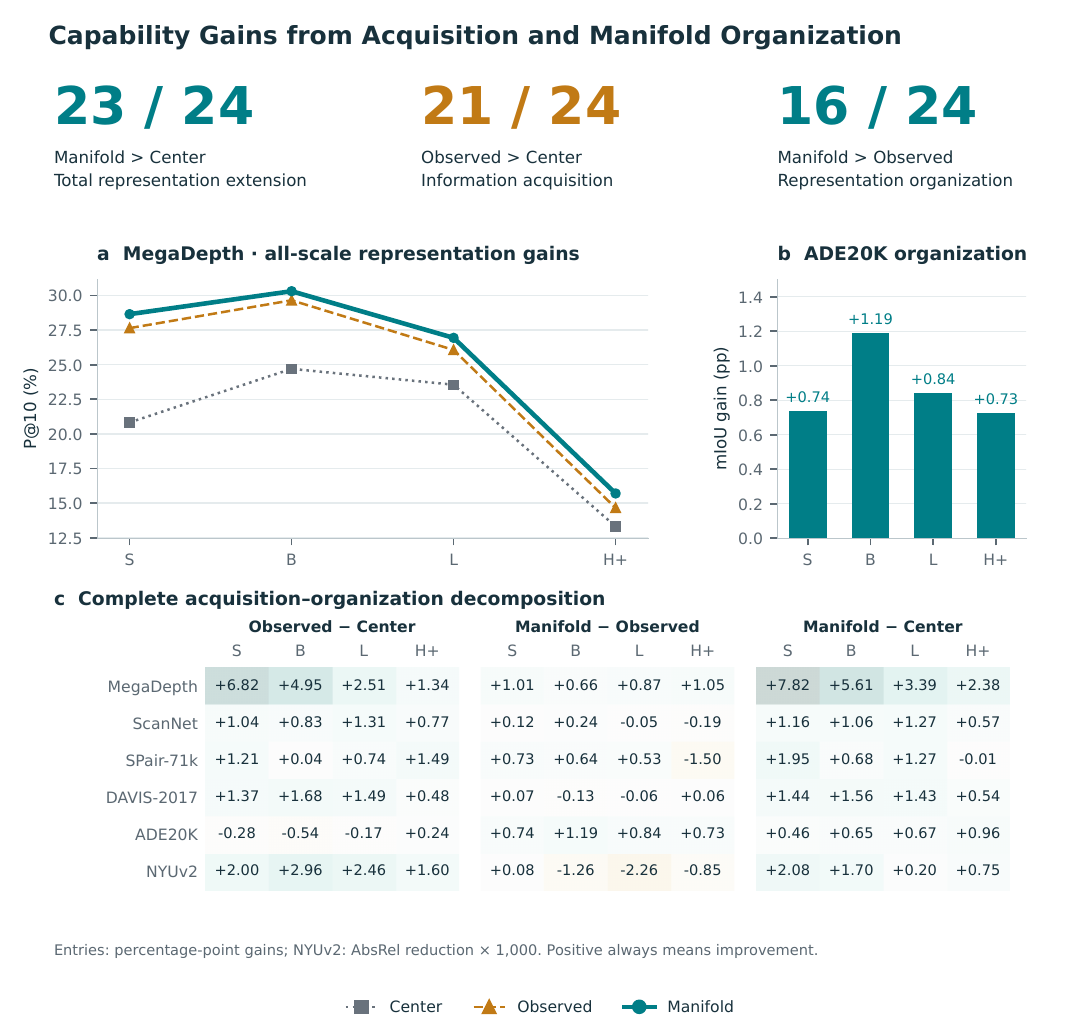}}
  \caption{\textbf{Capability gains from acquisition and Manifold organization.} \Observed exceeds \Center in 21 of 24 task-scale conditions, \Manifold exceeds \Center in 23, and \Manifold exceeds \Observed in 16. The full matrix separates Information Acquisition Value, Representation Organization Value, and Total Representation Extension in task-native metrics.}
  \label{fig:capabilities}
\end{figure}

The 72 primary scores are reported in \tabref{tab:downstream}, and the three controlled differences are summarized in \figref{fig:capabilities}. \Observed exceeds \Center in 21 of 24 task-scale conditions. \Manifold exceeds \Center in 23. The canonical interface therefore omits information that downstream tasks can use, even though the backbone and task reader remain fixed or tightly matched.

\begin{table}[H]
  \centering
  \PMTTcenterasset{\includegraphics[width=\PMTTAssetWidth]{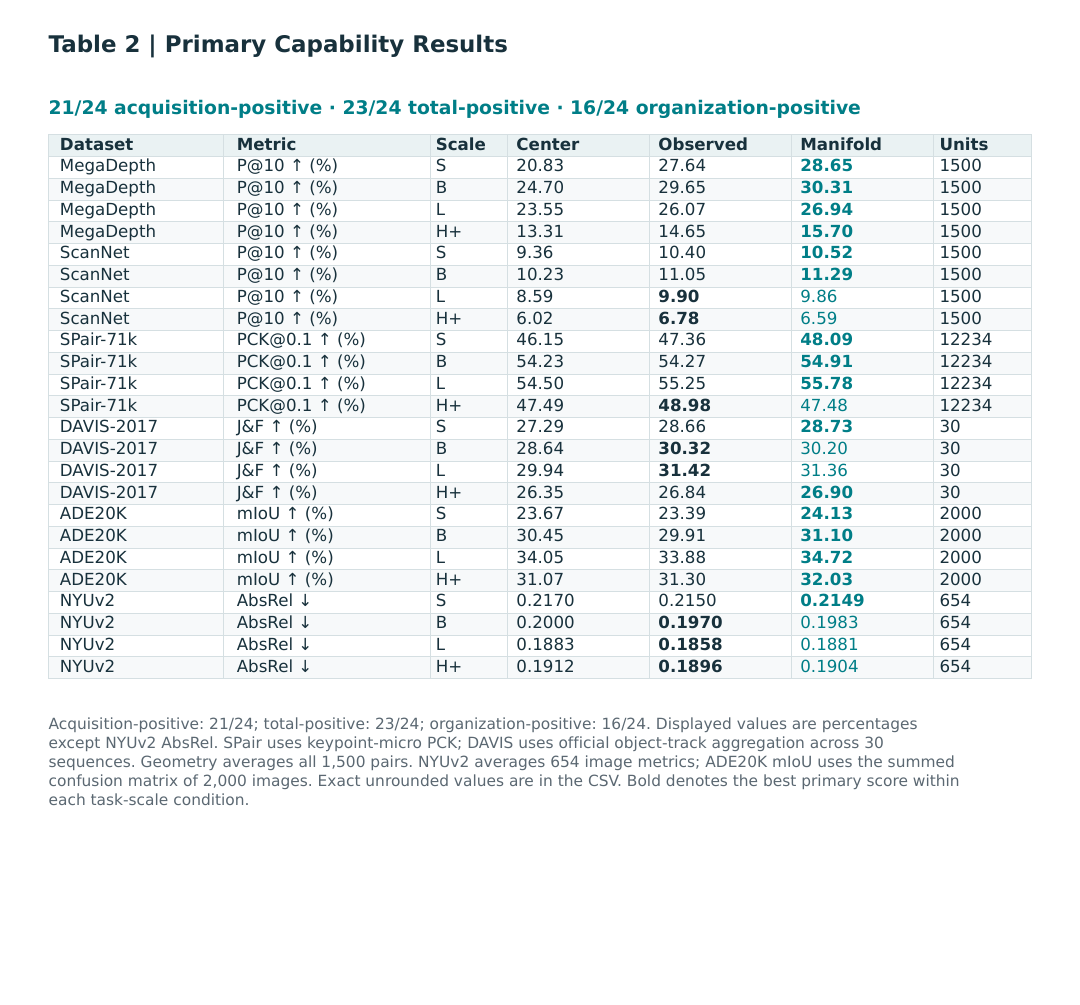}}
  \caption{\textbf{Primary capability results.} Displayed values are percentages except NYUv2 AbsRel. Bold marks the best primary representation within each task-scale condition. Controls and supplementary pose conditions do not enter the 72-cell matrix.}
  \label{tab:downstream}
\end{table}
\FloatBarrier

\subsection{Physical correspondence gives the clearest current consequence}
MegaDepth shows the strongest hierarchy. At S, P@10 increases from 20.83 for \Center to 27.64 for \Observed and 28.65 for \Manifold. The same ordering holds at B, L, and H+. Relative to \Center, the full-set mean gains are 7.82, 5.61, 3.39, and 2.38 percentage points. The organization gain remains positive at every scale.

ScanNet shows broad acquisition and total-extension gains in indoor geometry. \Manifold is best at S and B, while \Observed is slightly higher at L and H+. The two physical correspondence datasets therefore agree that additional same-patch states improve matching, and MegaDepth further shows a consistent benefit from continuous organization.

\begin{figure}[H]
  \centering
  \PMTTcenterasset{\includegraphics[height=\PMTTFigureFiveHeight,keepaspectratio]{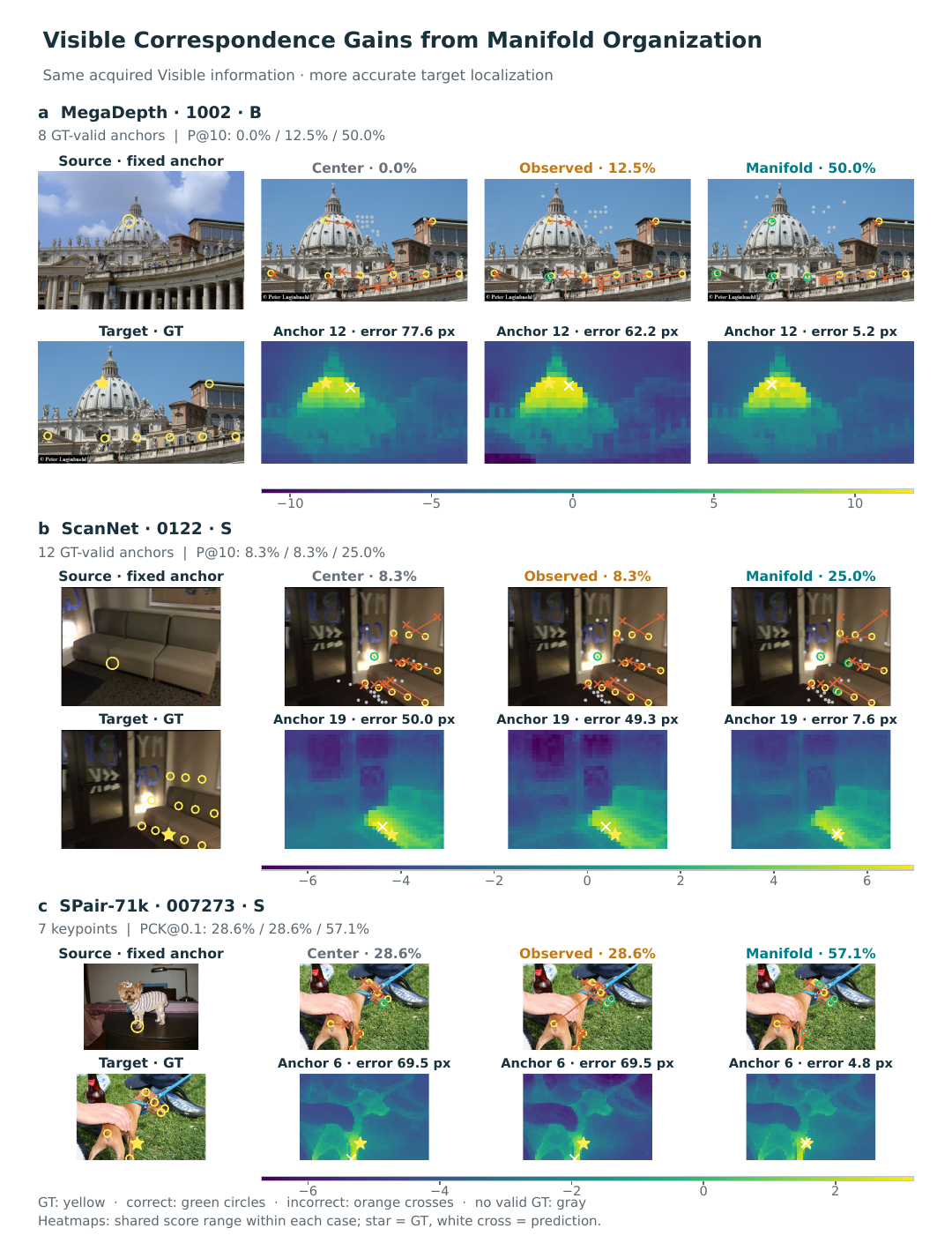}}
  \caption{\textbf{Visible correspondence gains from Manifold organization.} The largest case shows outdoor MegaDepth correspondence, followed by indoor ScanNet and semantic SPair correspondence. Each case improves over both \Center and \Observed. Score fields use a shared within-case normalization, so localization sharpness and peak position are directly comparable. The examples visualize the mechanism, while \figref{fig:capabilities} and \tabref{tab:downstream} report complete outcomes.}
  \label{fig:qualitative}
\end{figure}

The real correspondence cases in \figref{fig:qualitative} reveal the mechanism. \Manifold concentrates target score fields near the correct location and reduces correspondence error while using the same \Visible information as \Observed. The gain is not produced by a more expressive matcher.

\FloatBarrier

\subsection{Organization can recover value that direct aggregation loses}
ADE20K gives a complementary result. \Observed is below \Center at S, B, and L, showing that additional states are not automatically useful when they are aggregated without an appropriate organization. \Manifold is best at all four scales, with mIoU gains over \Center of 0.46, 0.65, 0.67, and 0.96 percentage points. The same information that can hurt as a discrete collection becomes useful when organized by its local geometry.

This is the clearest downstream demonstration of the acquisition-organization decomposition. More observations and better organization answer different questions. A representation can contain useful additional information while a particular readout of the raw states fails to exploit it.

\subsection{Different capabilities read different projections of the space}
The representation preference is task-dependent. SPair improves with \Manifold at S, B, and L, while H+ favors \Observed and leaves \Manifold near \Center. DAVIS gains mainly from acquisition, with small organization differences. NYUv2 improves modestly over \Center, but \Observed is usually the best organization. These outcomes do not weaken the broader representation result. They show that \PLFS contains multiple task-relevant projections and that a single linear Manifold organization does not preserve every useful statistic equally.

The current downstream conclusions belong to this analytic realization, its observation set, rank, and minimum-sufficient readers. The intrinsic evidence establishes a structured feature space. The task results show how one clean organization of that space behaves across six capabilities.

\section{Discussion and Open Research Space}
\paragraph{The patch token is an interface, not a complete measurement of representational content.}
A canonical token remains the most natural center of a patch representation. \PMTT does not remove that center. It shows that the center can be an incomplete readout of what the pretrained backbone can express about the patch. The distinction between backbone content and interface exposure matters: a model may already contain useful information that downstream systems never receive because the standard interface returns one state.

\paragraph{Observation design becomes part of representation design.}
Dense Orbital Views are not merely more test-time crops. They define which states of the patch representation are exposed. The identity constraint prevents the representation from becoming an arbitrary neighborhood, while dense directional sampling turns observation changes into a coherent local space. Future backbones may learn to expose this space more efficiently, but the scientific object remains meaningful even when its acquisition mechanism changes.

\paragraph{Variation can be information.}
Much of self-supervised learning treats view variation as a nuisance to remove or a relation used only during training. The intrinsic results show a different regime. Observation-induced differences among same-patch states are organized, patch-specific, and predictive. Collapsing them to one invariant point can discard useful structure. The appropriate goal may be selective invariance: stable identity with an explicit representation of lawful local variation.

\paragraph{Manifold is a strong origin-point organization, not a closed endpoint.}
The analytic \Manifold demonstrates that a simple centered low-dimensional object can read structure already present in a frozen backbone. Its strong intrinsic advantage and broad downstream gains make it a central contribution of the current work. Its task selectivity also opens the next questions. Nonlinear organization, learned local coordinates, asymmetric or task-conditioned projections, and compact state-family codes may preserve additional information that the current symmetric rank-8 factor does not.

\paragraph{A direct route to future learning objectives.}
Visible-to-Hidden provides an immediate self-supervised target: predict withheld same-patch states from visible orbital states. This objective differs from matching two global views or reconstructing masked pixels. It asks a model to learn the geometry of one patch's observation-conditioned state family. Related directions include learned acquisition, adaptive observation budgets, pretraining with explicit patch-level local objectives, backbone-native state-family outputs, and extension to video, real multi-view imagery, and 3D observations.

The present intrinsic laws are established on DINOv3 S/B/L/H+ with two independent cohorts. Broader universality across model families and training paradigms remains an empirical question. The downstream values likewise describe the current clean analytic realization. These boundaries define the next experiments without reducing the scientific update already established here.

\section{Conclusion}
\PMTT changes the representation object associated with a ViT patch. A canonical token remains the center, while identity-preserving observations expose a Representation State Family and open \PLFS. Dense Orbital Views acquire this information, Visible-to-Hidden tests its predictive structure, intrinsic experiments reveal its low-dimensional patch-specific laws, and \Manifold provides a continuous organization that is substantially stronger than direct observed states. A frozen zero-post-training realization produces broad capability gains across correspondence, propagation, depth, and semantics. The resulting conclusion is simple: a patch is more than the token returned by one canonical observation, and pretrained visual backbones already contain a richer patch representation than the standard interface exposes.

\FloatBarrier
\pdfbookmark[0]{References}{references}
\bibliographystyle{plainnat}
\bibliography{references}

\FloatBarrier
\clearpage
\appendix
\pdfbookmark[0]{Appendix}{appendix}
\begin{center}
{\Large\bfseries\color{pmttblue} Appendix\par}
\vspace{6pt}
{\color{pmttgray!45}\rule{0.34\textwidth}{0.5pt}\par}
\end{center}
\vspace{10pt}
\section{Definitions and Analytic Organization}
\label{app:definitions}
\subsection{State normalization and tangent variation}
All intrinsic geometry is formed around the unit canonical center $\bm{u}=\bm{c}/\|\bm{c}\|_2$, while the raw center norm is retained for downstream readers that use it. Observation states are normalized before local geometry is constructed. The tangent residual
\begin{equation}
  \bm{r}_j=(\bm{v}_j-\bm{u})-\bm{u}\bm{u}^{\top}(\bm{v}_j-\bm{u})
\end{equation}
removes the center-radial component. The residual matrix is decomposed by SVD, and the leading eight scaled directions form $A$. Because $A\bm{z}$ is unchanged under joint orthogonal rotation of the columns and coefficients, factor comparisons use spectra and subspaces rather than signed basis vectors.

\subsection{Bounded projection and coverage}
For a Hidden state $\bm{h}$, Manifold distance solves
\begin{equation}
  d_{\mathcal{M}}^2(\bm{h})=\min_{\|\bm{z}\|_2\leq1}\|\bm{h}-\bm{u}-A\bm{z}\|_2^2.
\end{equation}
The unbounded subspace distance removes the unit-ball constraint and is reported only as a diagnostic. \Observed distance is the minimum squared distance to the canonical and 16 Visible points. Pooled coverage uses summed errors over all Hidden states in the cohort, while anchor-mean coverage first computes coverage per anchor and then averages. The two quantities answer different weighting questions and remain separate.

\subsection{Spectrum, effective rank, energy, and overlap}
Let $\sigma_i$ be singular values and $p_i=\sigma_i^2/\sum_j\sigma_j^2$. Effective rank is the participation ratio
\begin{equation}
  r_{\mathrm{eff}}=\frac{1}{\sum_i p_i^2}.
\end{equation}
Top-$k$ energy is $\sum_{i=1}^k p_i$. Subspace overlap compares Visible-derived factors with the Full-32 Diagnostic Ceiling using basis-invariant principal-angle quantities at ranks 1, 2, 4, and 8.

\section{Realization and Evaluation Details}
\label{app:realization}
\subsection{Backbone and numerical configuration}
DINOv3 S/B/L/H+ use widths 384/768/1024/1280, blocks 12/12/24/32, heads 6/12/16/20, patch size 16, and the official final normalization. Structured downstream inference batches the canonical and 16 Visible views. Intrinsic extraction uses the canonical forward and orbital chunks of 32. Backbone inference uses BF16 autocast, outputs and geometry use FP32, and TF32 is disabled.

\subsection{Reader families}
SPair and DAVIS use regularized scalar models over representation-native features. The Center basis contains canonical similarity, norm, polynomial, and hinge statistics. The Observed basis contains symmetric statistics across Visible similarities. The Manifold basis contains bounded and unbounded distances, explained ratios, coefficient norms, boundary excess, spectrum concentration, and orthogonal residuals. Coefficients are shared across backbone scales after scale-specific standardization.

For NYUv2, a scale-specific Center-only linear axis predicts log depth. Observed and Manifold use small shared scalar mixtures of the axis responses and geometry statistics. For ADE20K, a scale-specific Center-only 150-class linear classifier is shared by all representations. Observed and Manifold use class-independent scalar mixtures applied after the shared logits or class-axis radii are computed. This design keeps the task axis fixed while testing the representation supplied to it.

\subsection{Correspondence protocols}
SPair evaluates official source keypoints against a common target candidate grid. MegaDepth and ScanNet use 32 deterministic source anchors, a target grid with long side 640, one-to-one assignment, and fixed local refinement. The 64-anchor B/H+ pose experiment is supplementary and does not replace the primary 32-anchor correspondence matrix. Pose correctness thresholds are 0.5 resized pixels for MegaDepth and 4.0 for ScanNet.

\subsection{Evaluation substrate}
\setcounter{table}{0}
\renewcommand{\thetable}{A\arabic{table}}
\begin{table}[H]
  \centering
  \PMTTcenterasset{\includegraphics[width=\PMTTAssetWidth]{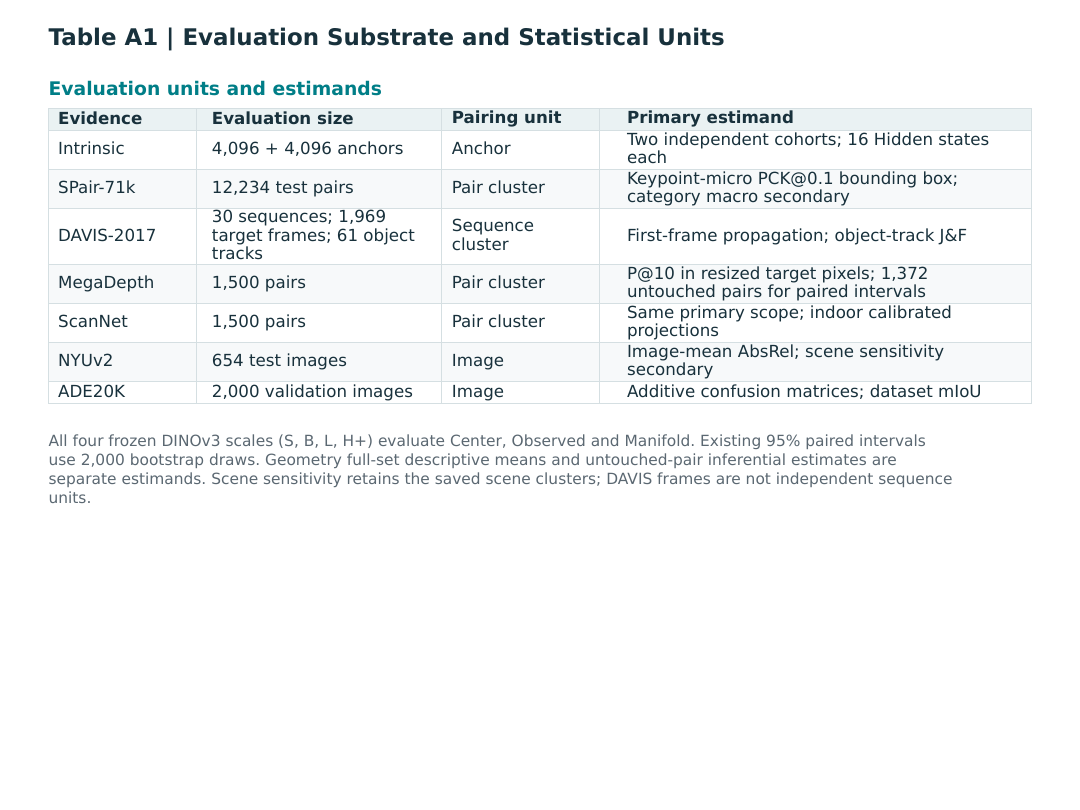}}
  \caption{\textbf{Evaluation substrate and statistical units.} The table records official units, primary metrics, paired units, uncertainty procedures, and control availability.}
  \label{tab:a1}
\end{table}

\begin{table}[H]
  \centering
  \PMTTcenterasset{\includegraphics[width=\PMTTAssetWidth]{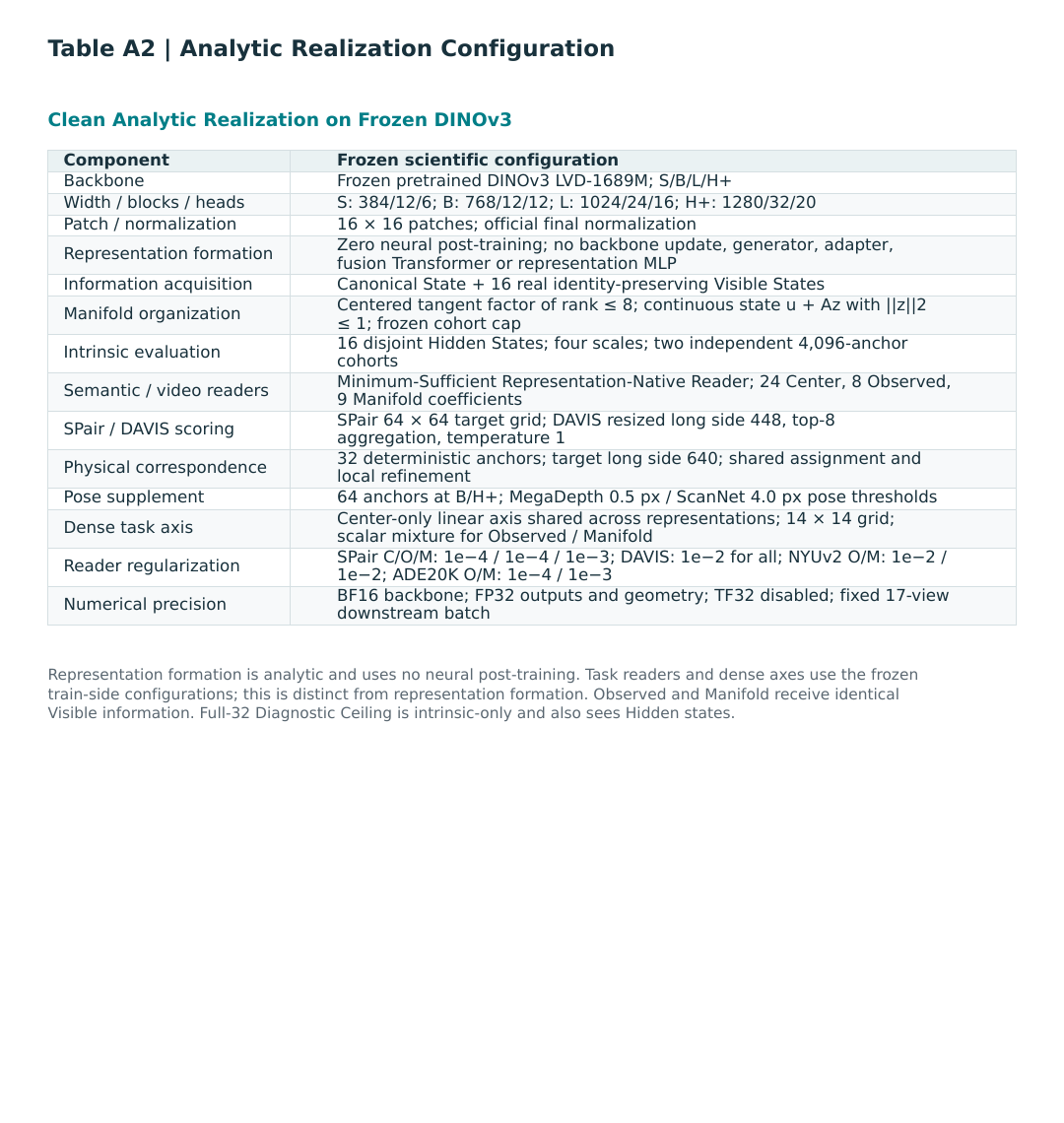}}
  \caption{\textbf{Analytic realization configuration.} Representation formation is analytic and uses no neural post-training. Task readers and dense axes use the fixed train-side configurations listed here.}
  \label{tab:a2}
\end{table}

\FloatBarrier
\section{Complete Intrinsic Evidence}
\label{app:intrinsic}
\setcounter{figure}{0}
\renewcommand{\thefigure}{A\arabic{figure}}
The main paper reports the predictive and scaling laws that define \PLFS. This appendix separates cohort replication, spectral structure, attribution controls, bounds, and real observation families.

\begin{figure}[H]
  \centering
  \PMTTcenterasset{\includegraphics[width=\PMTTAssetWidth]{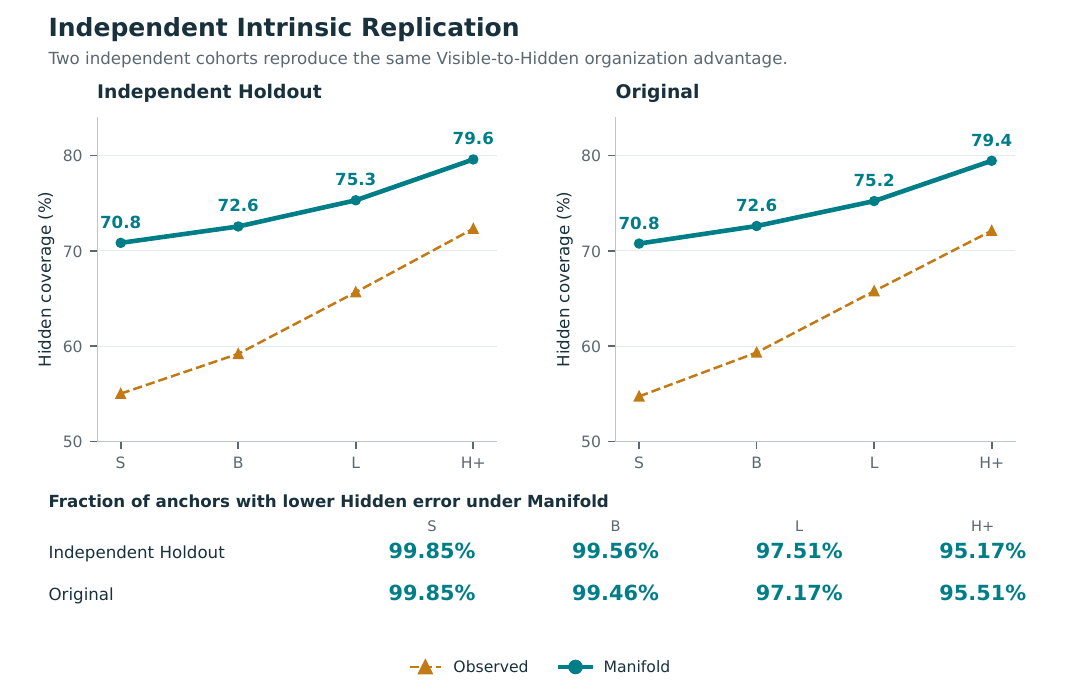}}
  \caption{\textbf{Independent intrinsic replication.} The Original Cohort and Independent Holdout Cohort each contain 4,096 anchors and independently reproduce the Manifold ordering at every scale.}
  \label{fig:a1}
\end{figure}

\begin{figure}[H]
  \centering
  \PMTTcenterasset{\includegraphics[width=\PMTTAssetWidth]{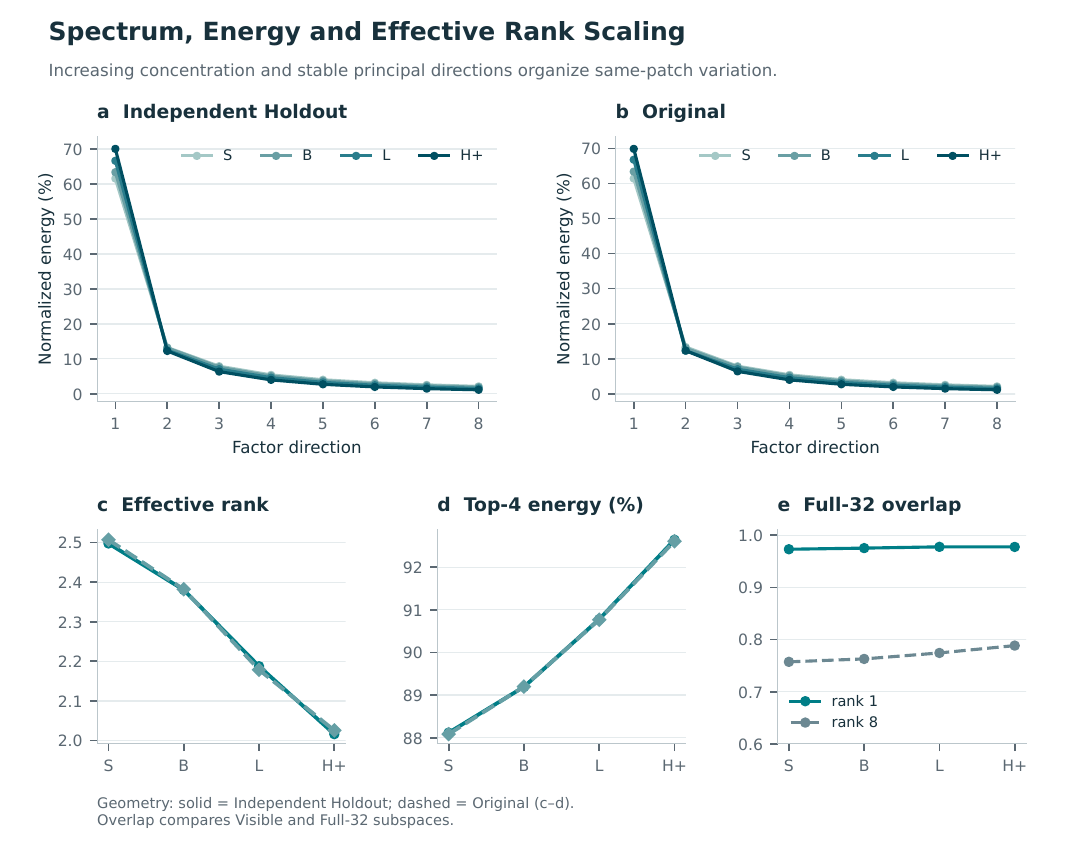}}
  \caption{\textbf{Spectrum, energy, and effective-rank scaling.} Mean normalized spectra, effective rank, Top-4 energy, and overlap with the Full-32 Diagnostic Ceiling are shown separately for both cohorts.}
  \label{fig:a2}
\end{figure}

\begin{figure}[H]
  \centering
  \PMTTcenterasset{\includegraphics[width=\PMTTAssetWidth]{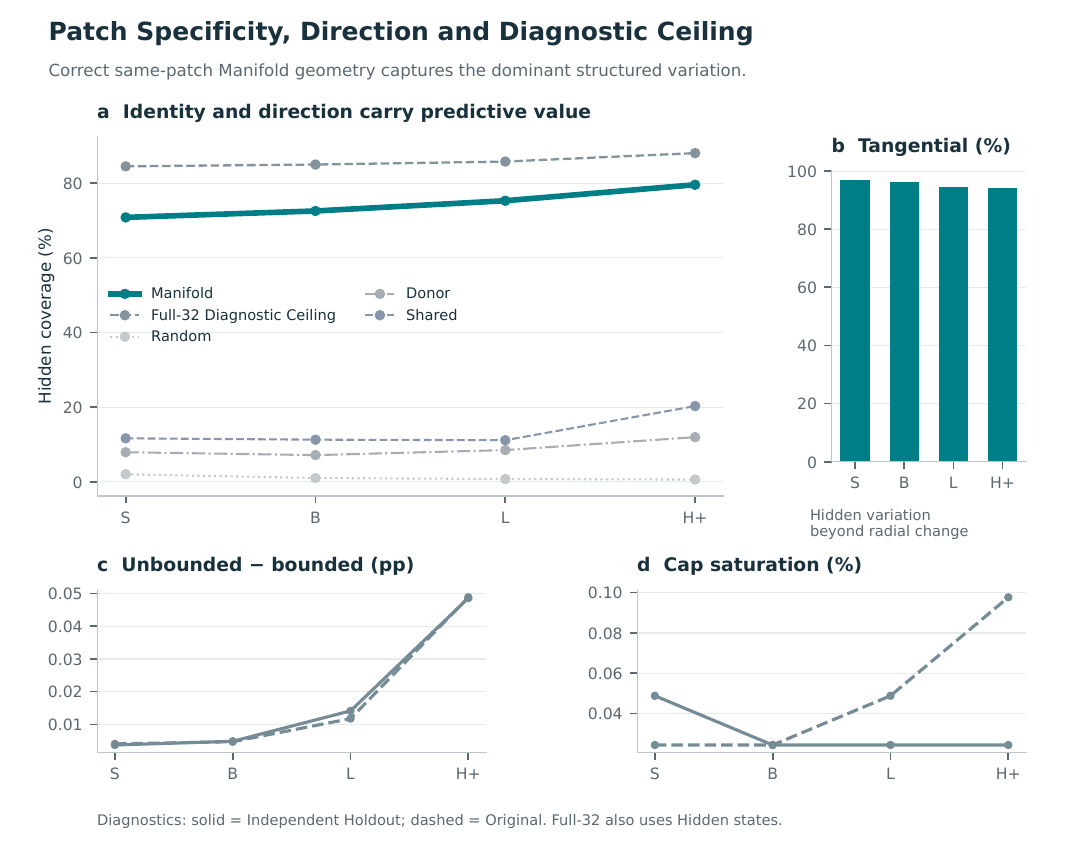}}
  \caption{\textbf{Patch specificity, direction, and diagnostic ceiling.} Correct same-patch geometry predicts substantially more Hidden variation than Random, Donor, or Shared geometry. Supporting panels report tangential variation, boundedness, cap saturation, and the Full-32 ceiling.}
  \label{fig:a3}
\end{figure}

\begin{figure}[H]
  \centering
  \PMTTcenterasset{\includegraphics[page=1,width=\PMTTAssetWidth]{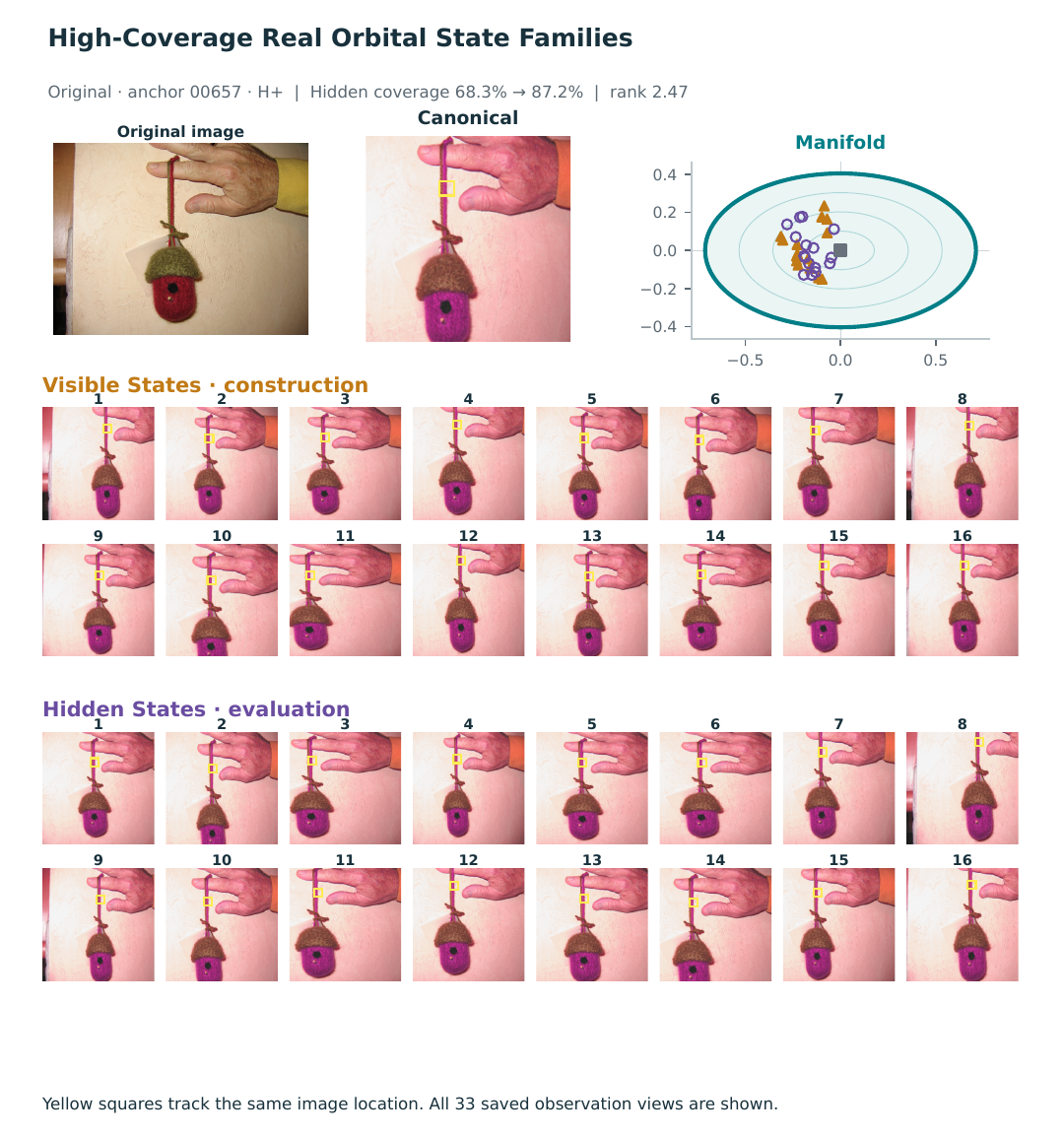}}
  \caption{\textbf{High-coverage real orbital state families, page 1 of 4.} Each page shows the original image, Canonical State, all 16 Visible and 16 Hidden observation crops, and the corresponding factor projection. Yellow squares locate the same underlying image point.}
  \label{fig:a4}
\end{figure}
\begin{figure}[H]\ContinuedFloat
  \centering
  \PMTTcenterasset{\includegraphics[page=2,width=\PMTTAssetWidth]{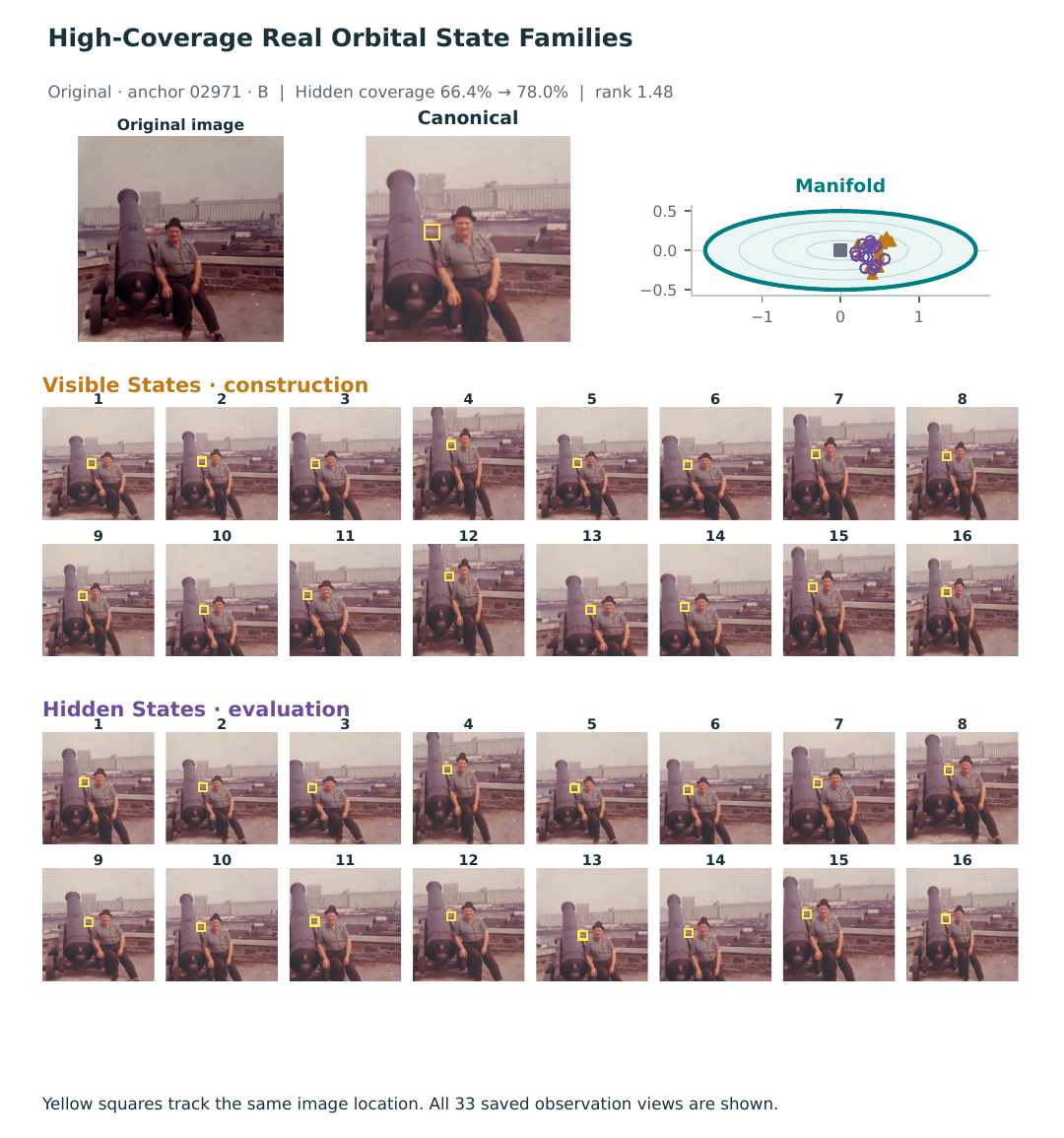}}
  \caption{\textbf{High-coverage real orbital state families, page 2 of 4.}}
\end{figure}
\begin{figure}[H]\ContinuedFloat
  \centering
  \PMTTcenterasset{\includegraphics[page=3,width=\PMTTAssetWidth]{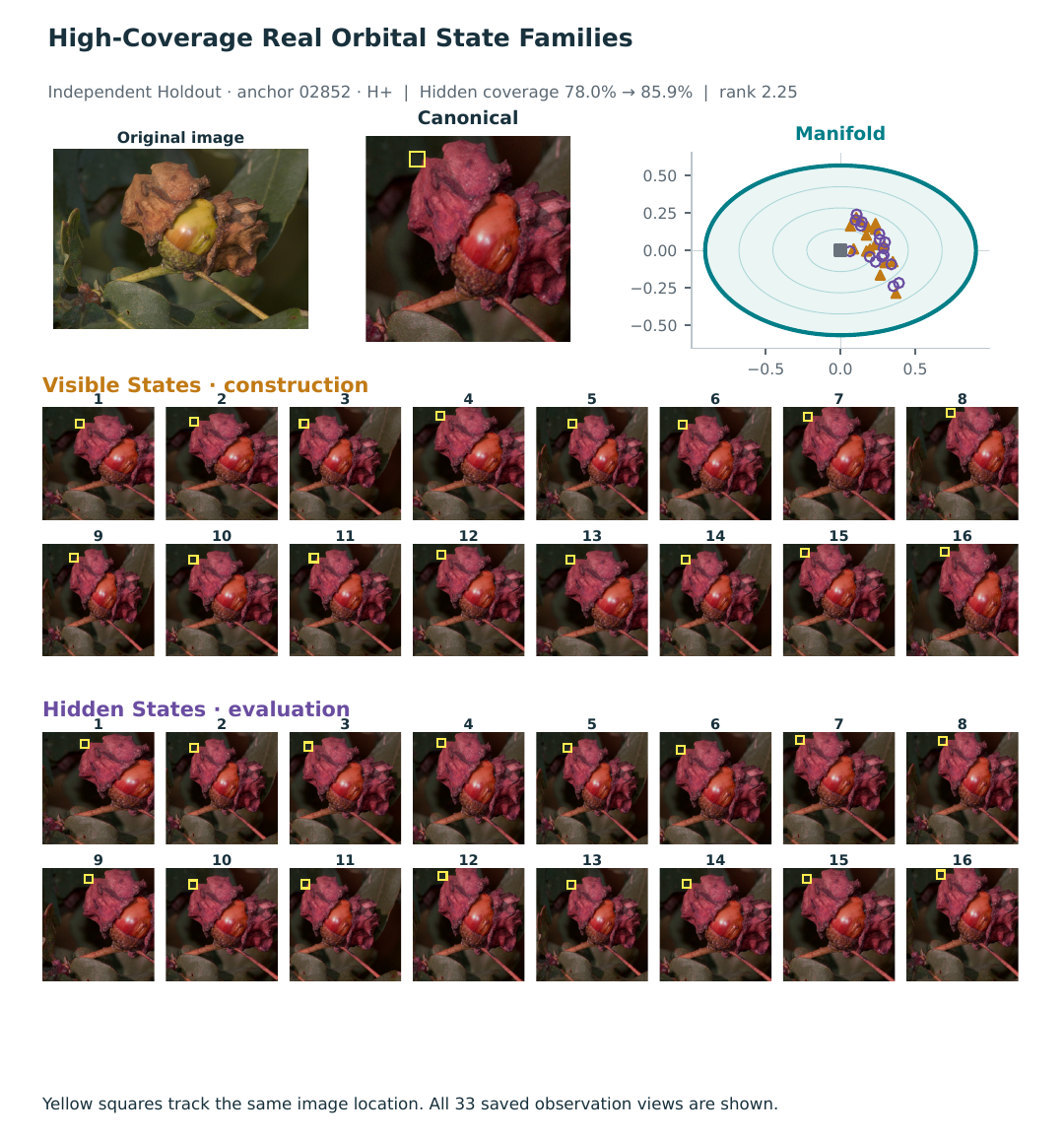}}
  \caption{\textbf{High-coverage real orbital state families, page 3 of 4.}}
\end{figure}
\begin{figure}[H]\ContinuedFloat
  \centering
  \PMTTcenterasset{\includegraphics[page=4,width=\PMTTAssetWidth]{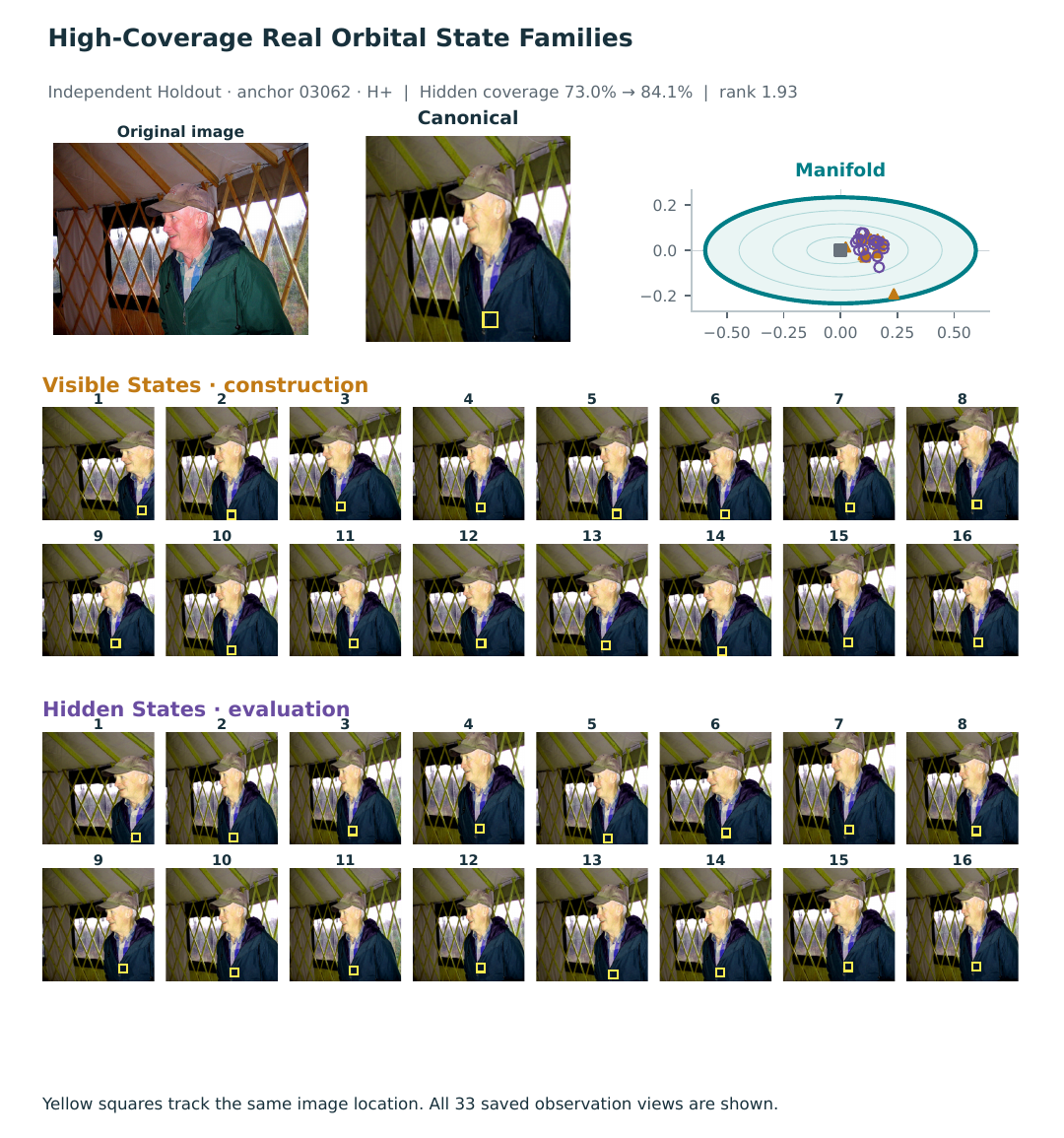}}
  \caption{\textbf{High-coverage real orbital state families, page 4 of 4.}}
\end{figure}

\begin{table}[H]
  \centering
  \PMTTcenterasset{\includegraphics[width=\PMTTAssetWidth]{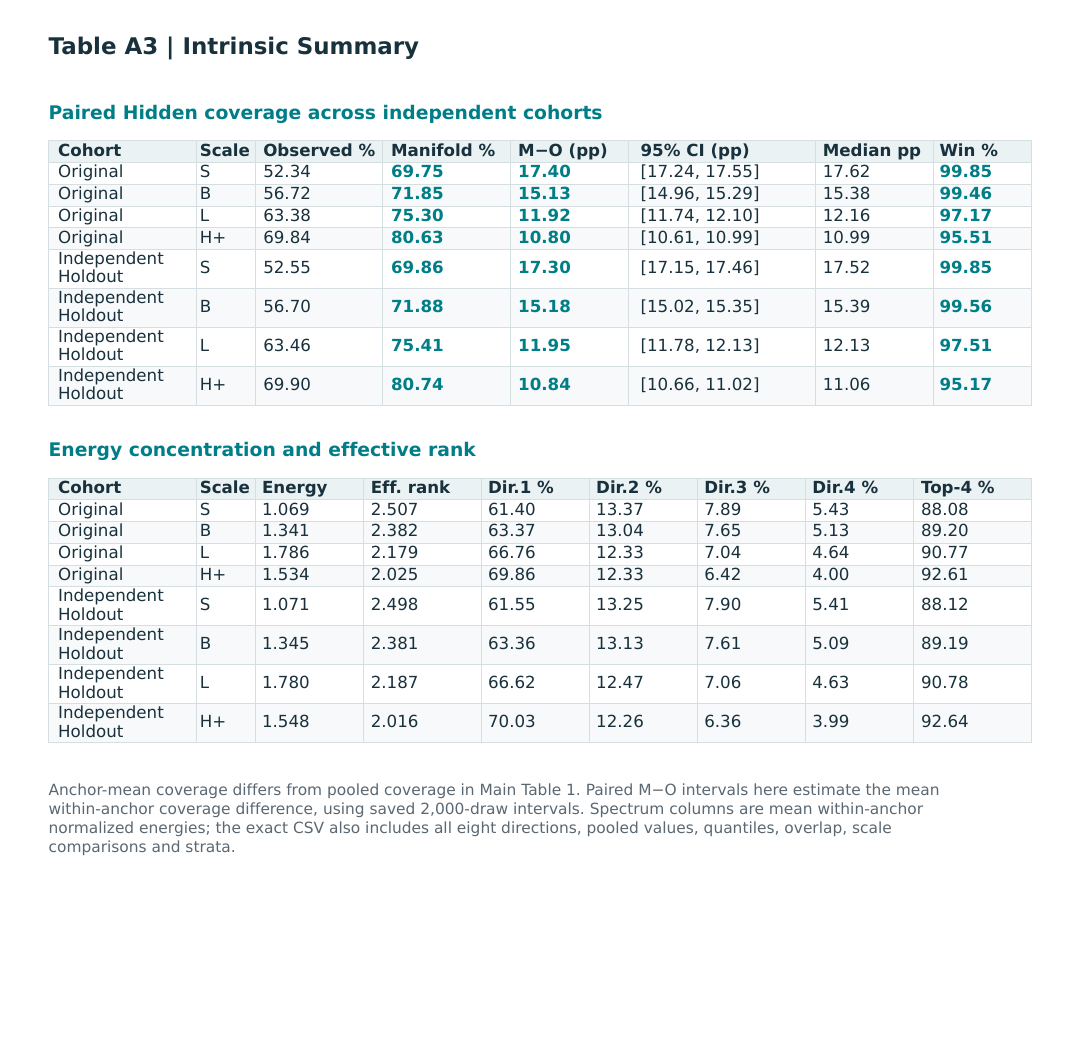}}
  \caption{\textbf{Complete intrinsic summary.} The table separates pooled and anchor-mean quantities, paired organization effects, spectra, overlap, and scale comparisons.}
  \label{tab:a3}
\end{table}

\begin{table}[H]
  \centering
  \PMTTcenterasset{\includegraphics[width=\PMTTAssetWidth]{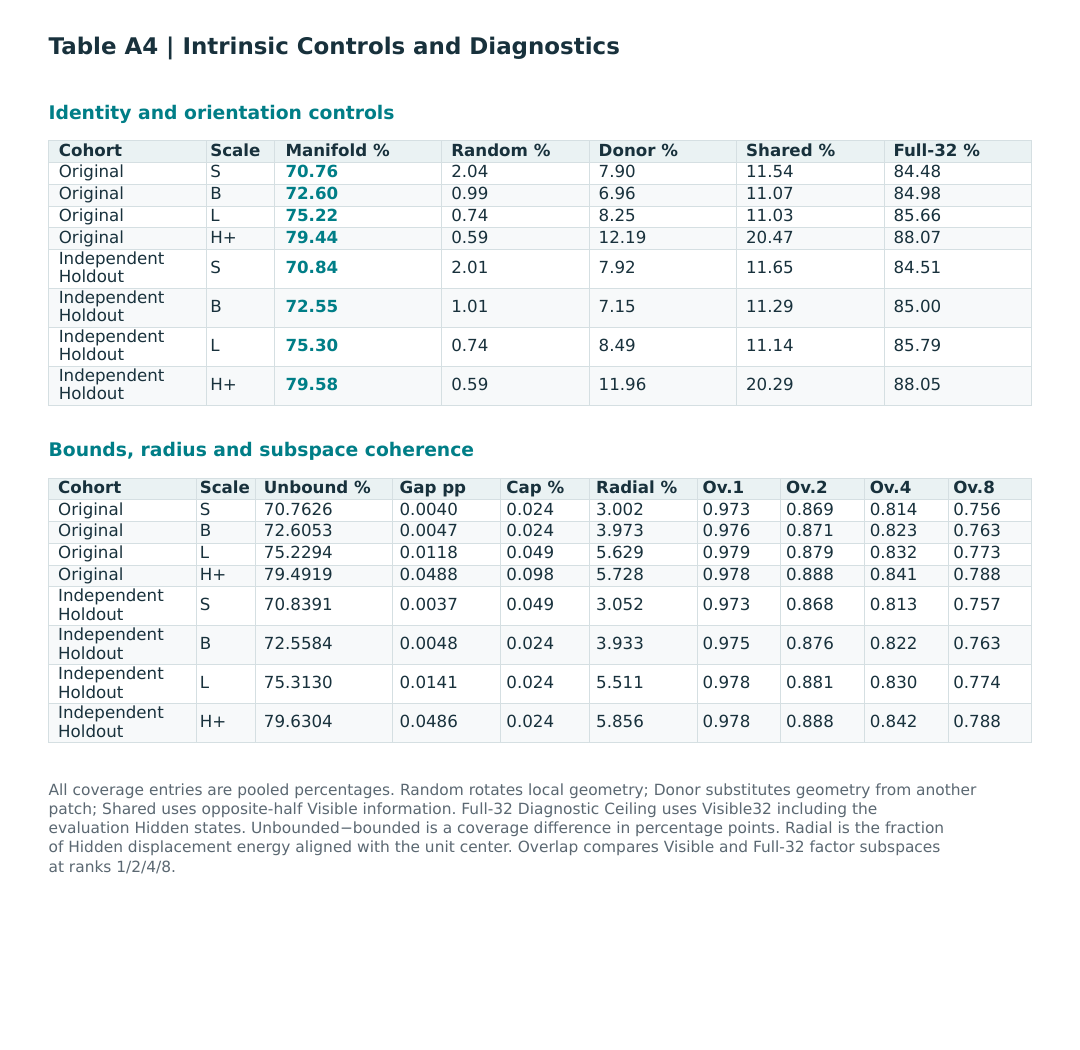}}
  \caption{\textbf{Intrinsic controls and diagnostics.} All coverage values are pooled percentages. Random, Donor, and Shared isolate direction and patch identity, while Full-32 provides a non-inference ceiling.}
  \label{tab:a4}
\end{table}

\FloatBarrier
\section{Complete Capability Evidence}
\label{app:capability}
\setcounter{figure}{0}
\renewcommand{\thefigure}{B\arabic{figure}}
\subsection{Paired effects and controls}
\begin{figure}[H]
  \centering
  \PMTTcenterasset{\includegraphics[height=\PMTTFigureBOneHeight,keepaspectratio]{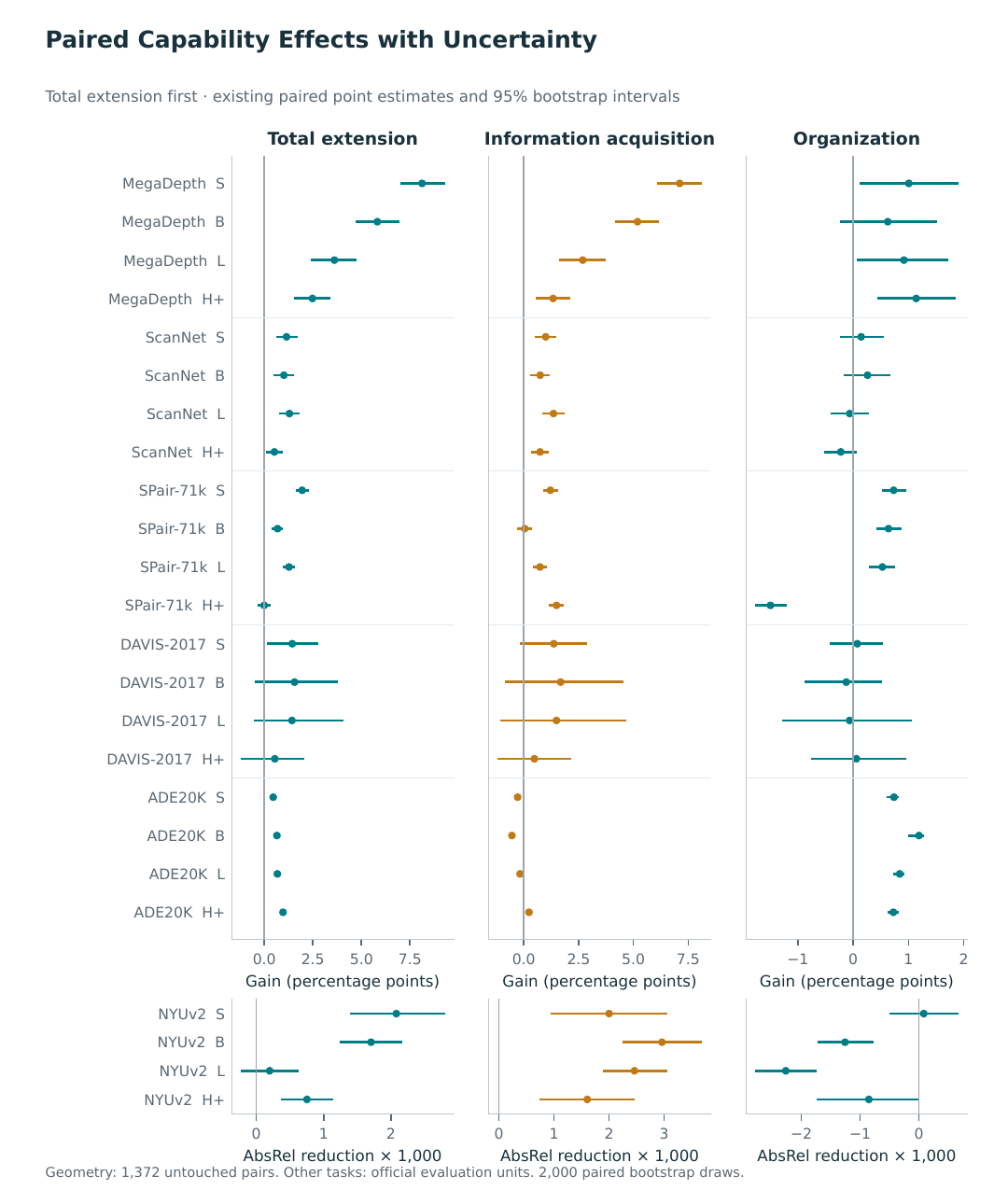}}
  \caption{\textbf{Paired capability effects with uncertainty.} Columns show Manifold$-$Center, Observed$-$Center, and Manifold$-$Observed with saved 95\% paired bootstrap intervals. NYUv2 uses AbsRel reduction in units of $10^{-3}$; other tasks use percentage points.}
  \label{fig:b1}
\end{figure}

\begin{figure}[H]
  \centering
  \PMTTcenterasset{\includegraphics[width=\PMTTAssetWidth]{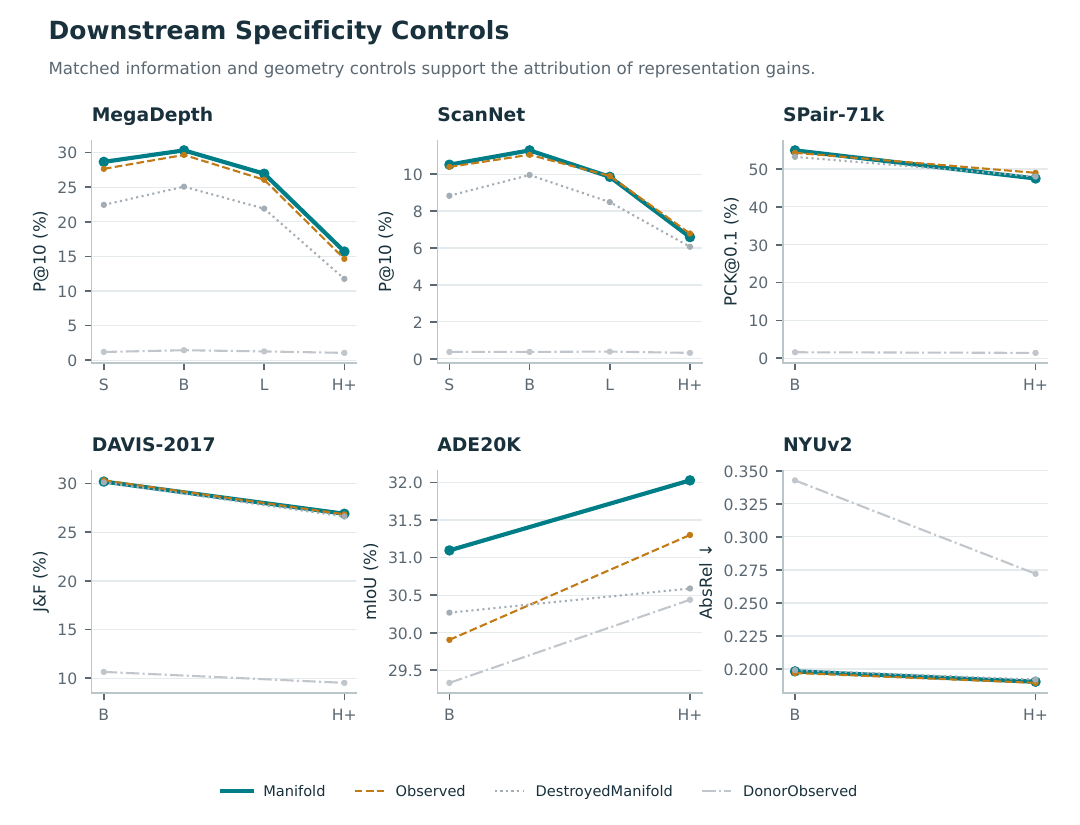}}
  \caption{\textbf{Downstream specificity controls.} DonorObserved substitutes donor Visible states. DestroyedManifold preserves the factor spectrum while destroying orientation. The primary reader is fixed.}
  \label{fig:b2}
\end{figure}

\subsection{Task-native secondary structure}
\begin{figure}[H]
  \centering
  \PMTTcenterasset{\includegraphics[width=\PMTTAssetWidth]{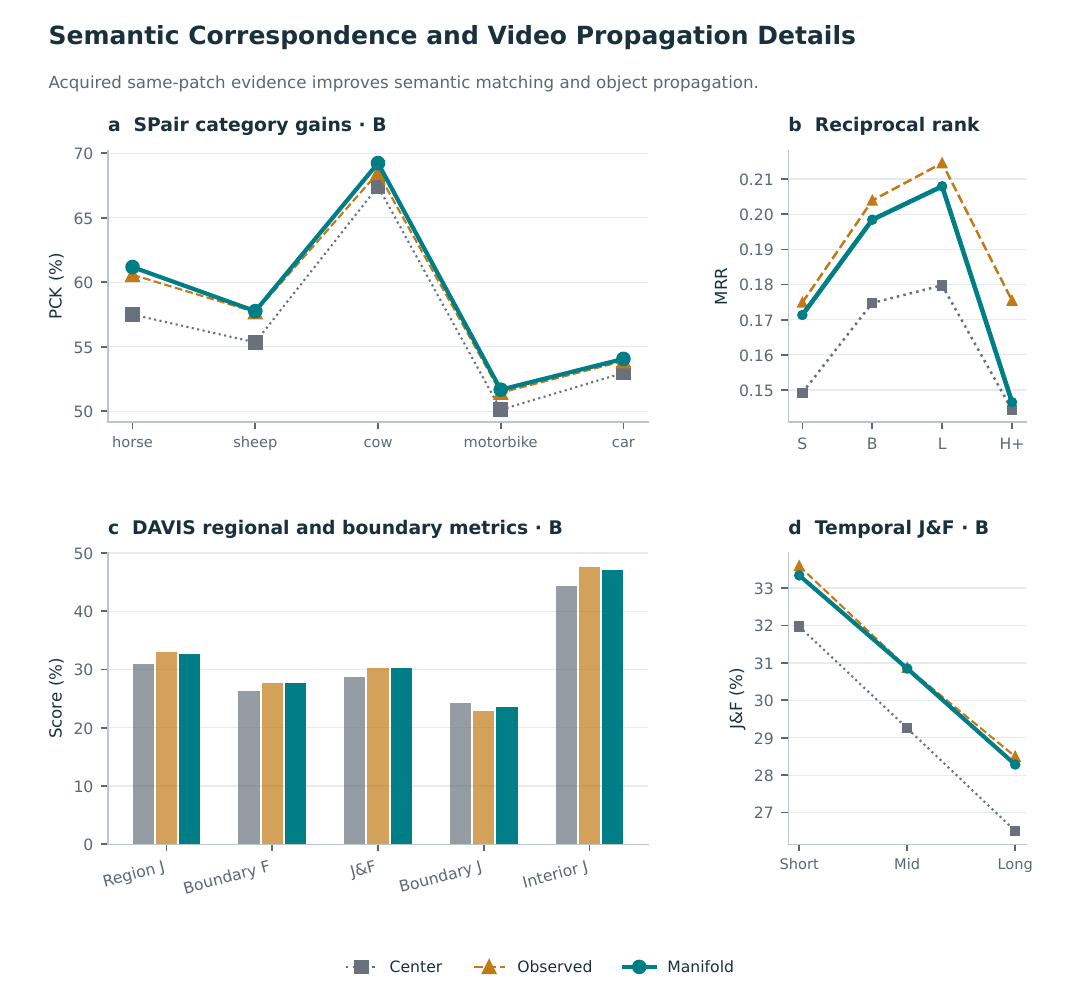}}
  \caption{\textbf{Semantic correspondence and video propagation details.} SPair category and rank structure are shown with DAVIS region, boundary, interior, and temporal metrics.}
  \label{fig:b3}
\end{figure}

\begin{figure}[H]
  \centering
  \PMTTcenterasset{\includegraphics[width=\PMTTAssetWidth]{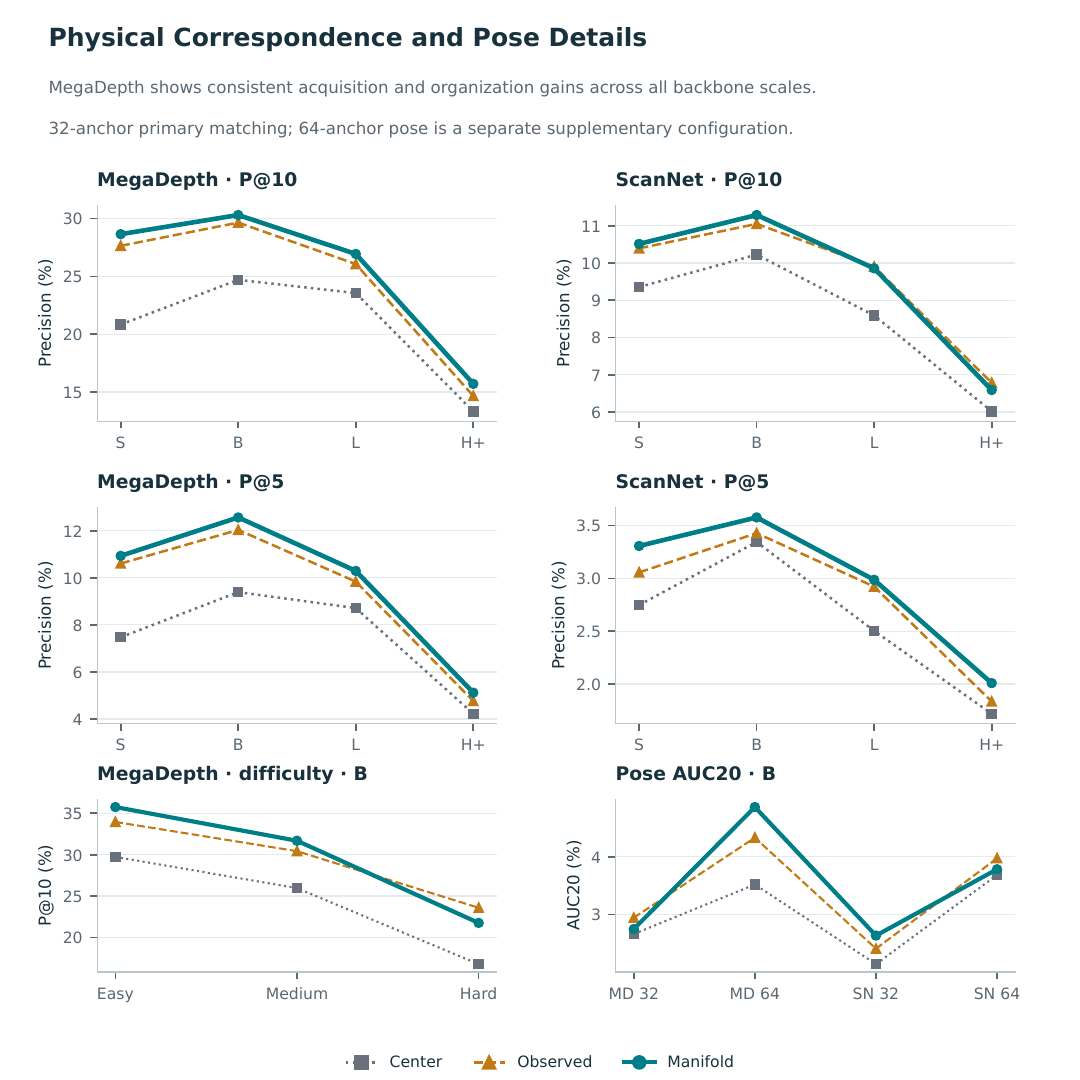}}
  \caption{\textbf{Physical correspondence and pose details.} Primary 32-anchor P@5/P@10 results, difficulty strata, pose AUC, failure, and the separate 64-anchor B/H+ pose supplement are reported.}
  \label{fig:b4}
\end{figure}

\begin{figure}[H]
  \centering
  \PMTTcenterasset{\includegraphics[width=\PMTTAssetWidth]{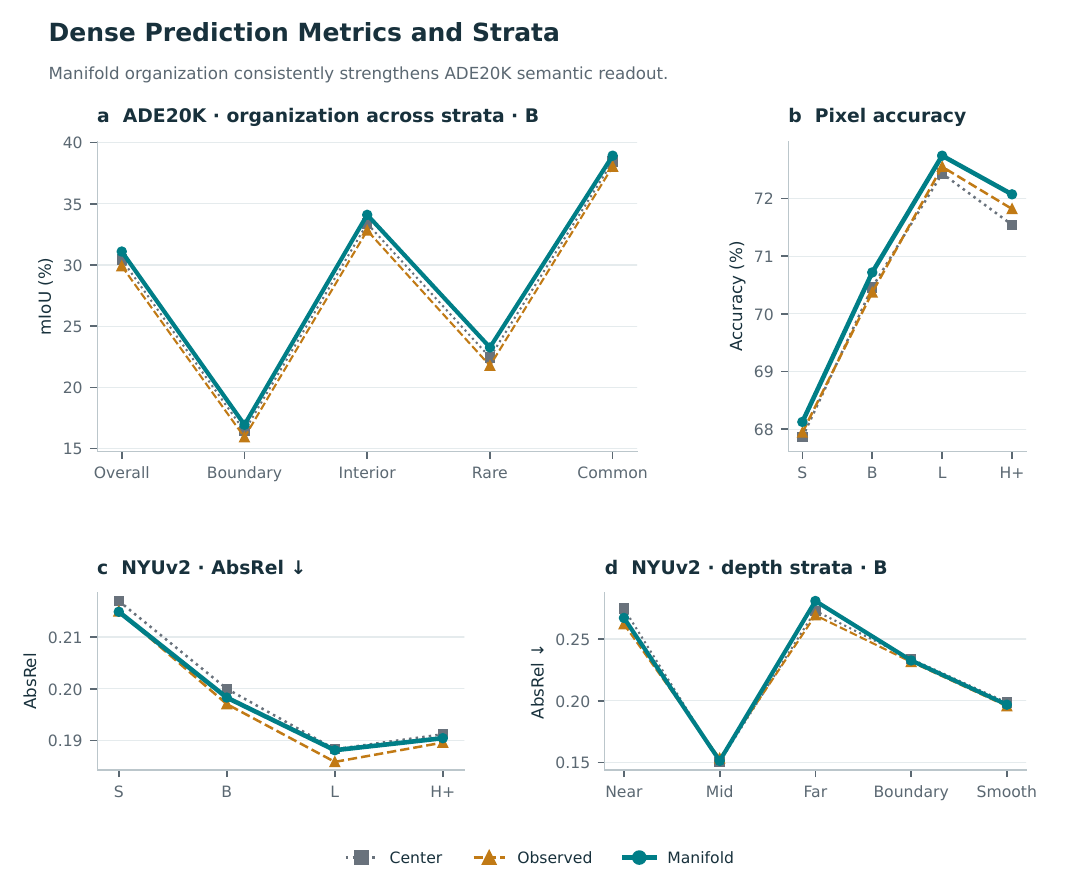}}
  \caption{\textbf{Dense prediction metrics and strata.} ADE20K semantic organization is shown with boundary, interior, rare, and common partitions. NYUv2 remains in native depth metrics and strata.}
  \label{fig:b5}
\end{figure}

\setcounter{table}{0}
\renewcommand{\thetable}{B\arabic{table}}
\begin{table}[H]
  \centering
  \PMTTcenterasset{\includegraphics[page=1,width=\PMTTAssetWidth]{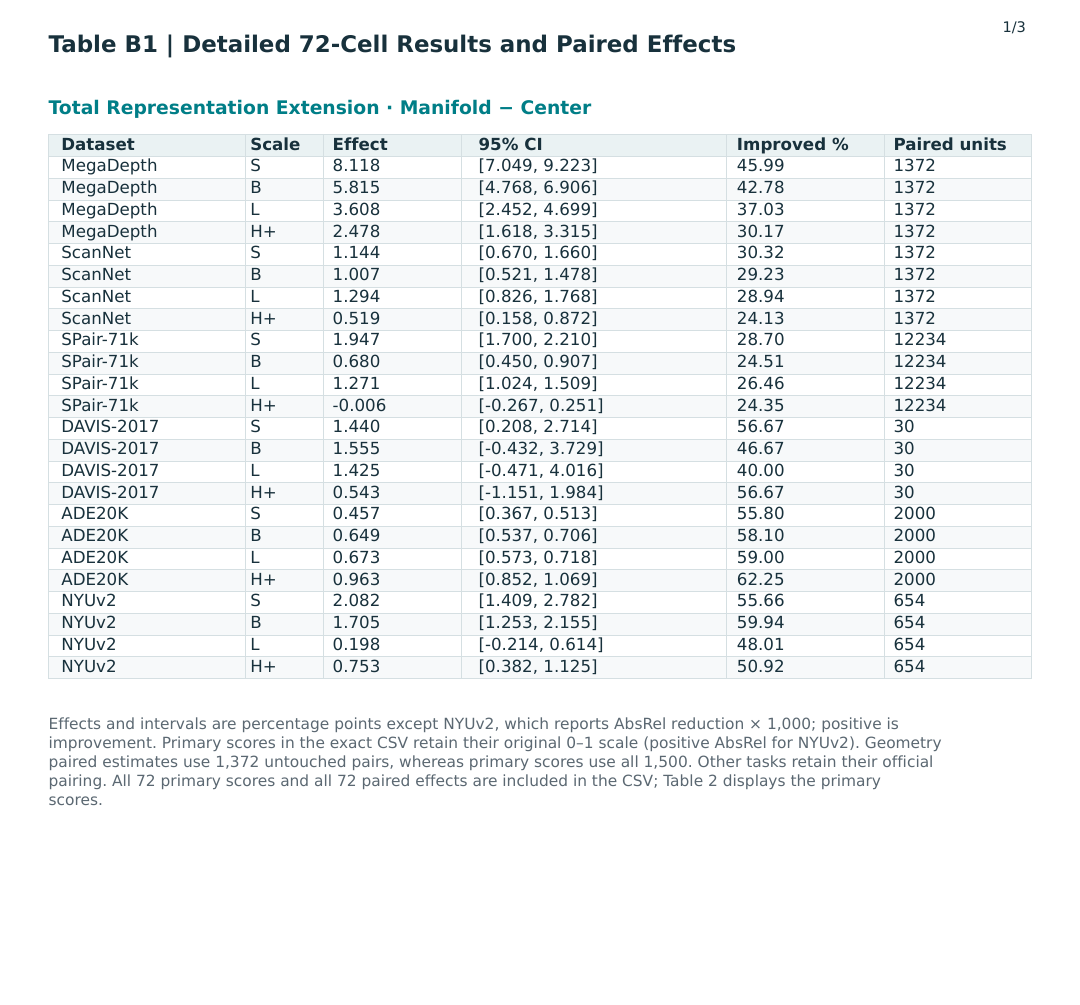}}
  \caption{\textbf{Detailed 72-cell results and paired effects, page 1 of 3.} All primary scores and all three paired effects are retained in the accompanying exact CSV.}
  \label{tab:b1}
\end{table}
\begin{table}[H]\ContinuedFloat
  \centering
  \PMTTcenterasset{\includegraphics[page=2,width=\PMTTAssetWidth]{tables/appendix/Table_B1.pdf}}
  \caption{\textbf{Detailed 72-cell results and paired effects, page 2 of 3.}}
\end{table}
\begin{table}[H]\ContinuedFloat
  \centering
  \PMTTcenterasset{\includegraphics[page=3,width=\PMTTAssetWidth]{tables/appendix/Table_B1.pdf}}
  \caption{\textbf{Detailed 72-cell results and paired effects, page 3 of 3.}}
\end{table}

\begin{table}[H]
  \centering
  \PMTTcenterasset{\includegraphics[width=\PMTTAssetWidth]{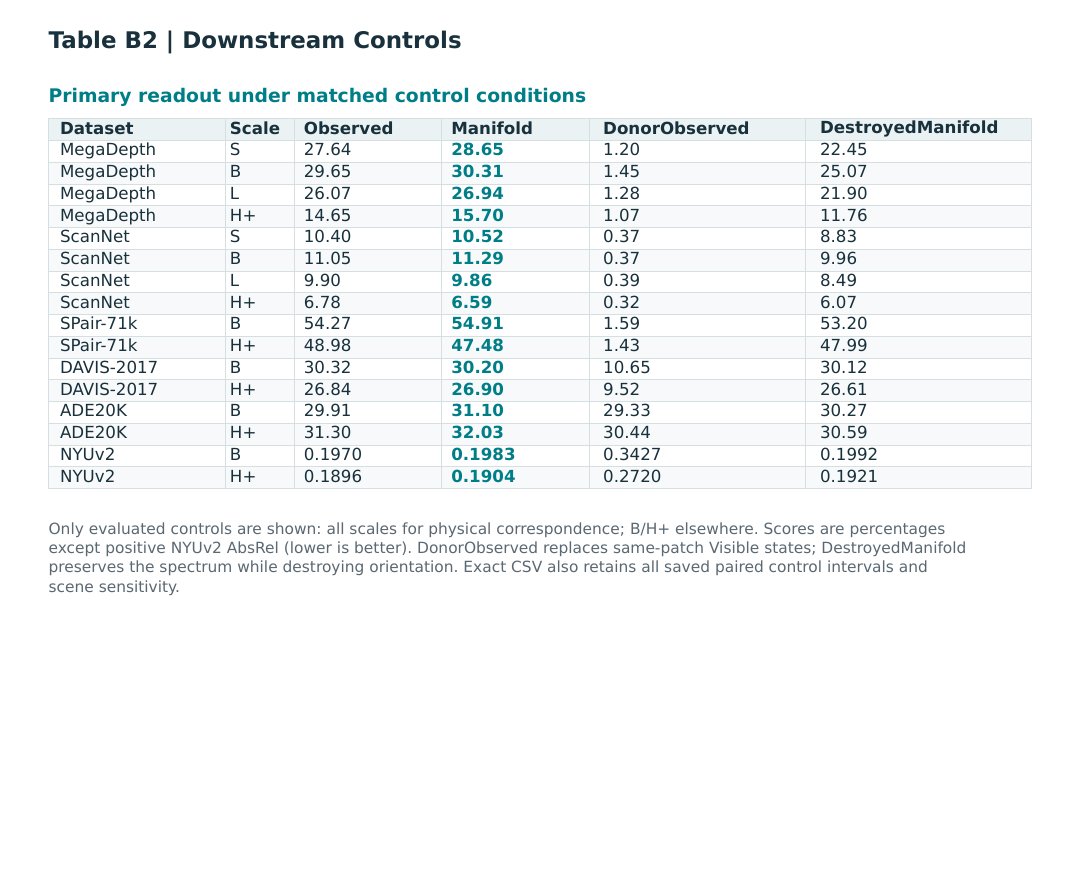}}
  \caption{\textbf{Downstream controls.} Only evaluated conditions are shown. Physical correspondence includes all scales; other capabilities include B and H+.}
  \label{tab:b2}
\end{table}

\begin{table}[H]
  \centering
  \PMTTcenterasset{\includegraphics[page=1,width=\PMTTAssetWidth]{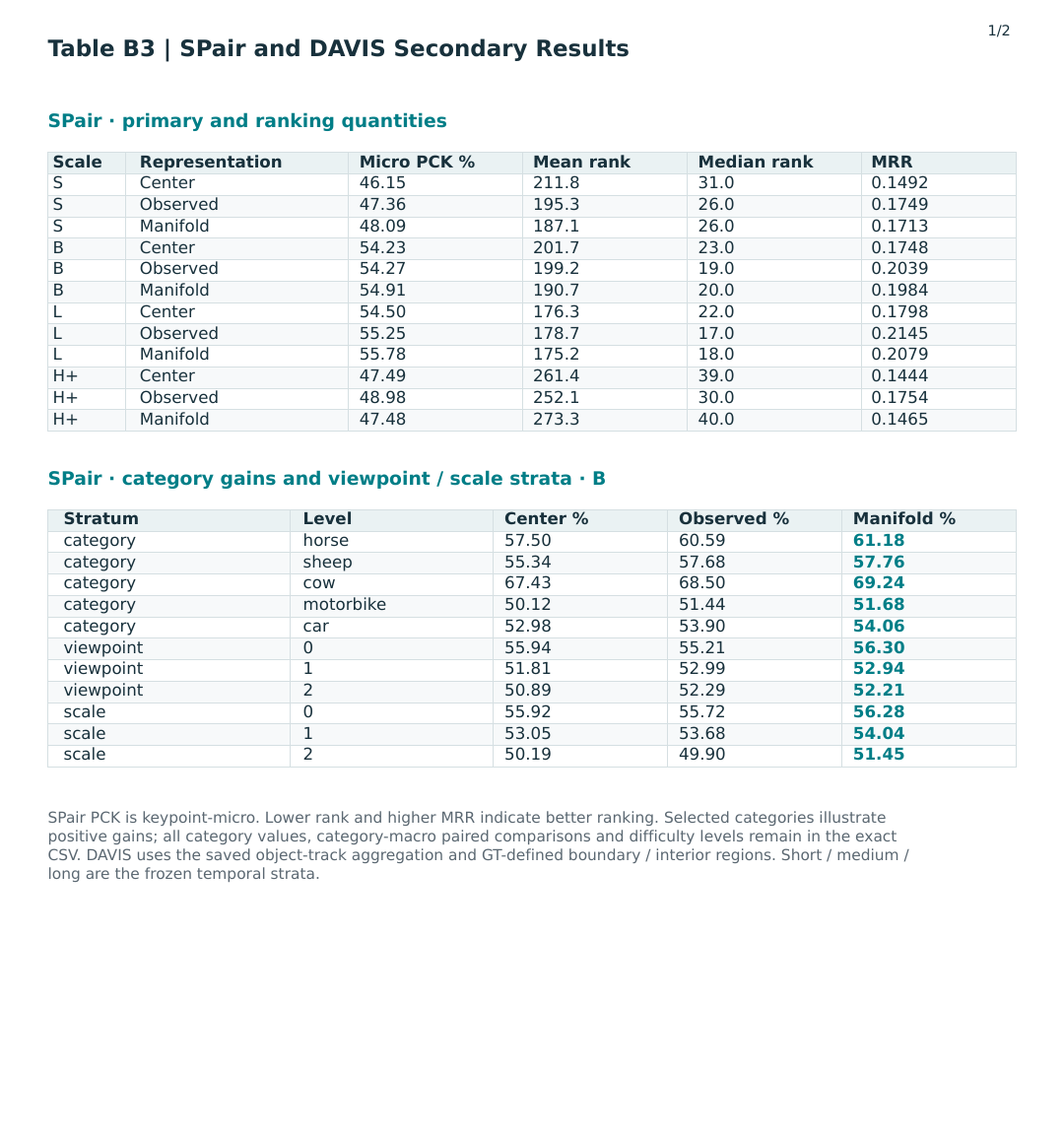}}
  \caption{\textbf{SPair and DAVIS secondary results, page 1 of 2.} Exact category, difficulty, rank, region, boundary, interior, and temporal results remain available in the CSV.}
  \label{tab:b3}
\end{table}
\begin{table}[H]\ContinuedFloat
  \centering
  \PMTTcenterasset{\includegraphics[page=2,width=\PMTTAssetWidth]{tables/appendix/Table_B3.pdf}}
  \caption{\textbf{SPair and DAVIS secondary results, page 2 of 2.}}
\end{table}

\begin{table}[H]
  \centering
  \PMTTcenterasset{\includegraphics[page=1,width=\PMTTAssetWidth]{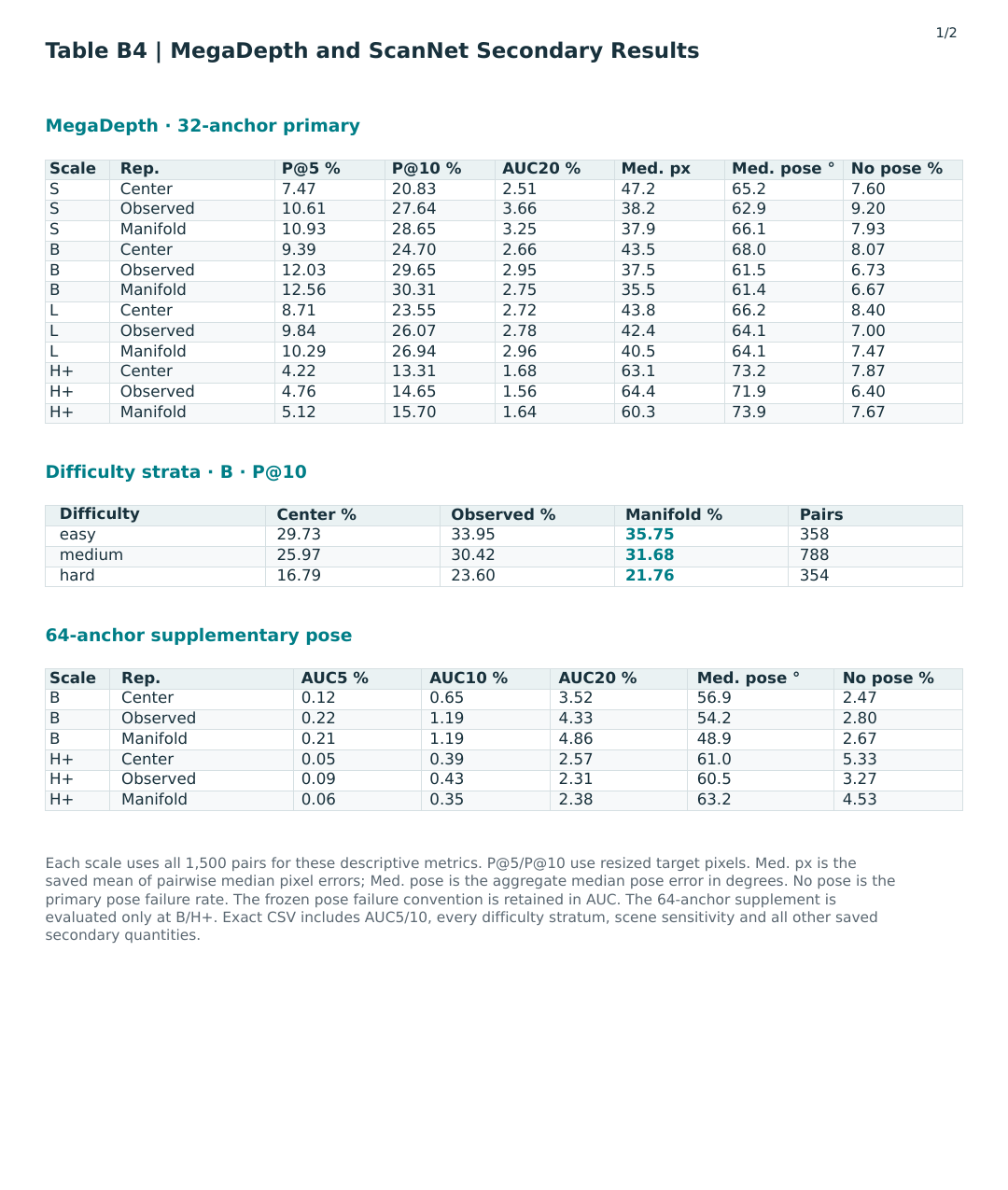}}
  \caption{\textbf{MegaDepth and ScanNet secondary results, page 1 of 2.} All descriptive metrics use the complete 1,500-pair sets; paired inference uses the fixed untouched subsets.}
  \label{tab:b4}
\end{table}
\begin{table}[H]\ContinuedFloat
  \centering
  \PMTTcenterasset{\includegraphics[page=2,width=\PMTTAssetWidth]{tables/appendix/Table_B4.pdf}}
  \caption{\textbf{MegaDepth and ScanNet secondary results, page 2 of 2.}}
\end{table}

\begin{table}[H]
  \centering
  \PMTTcenterasset{\includegraphics[width=\PMTTAssetWidth]{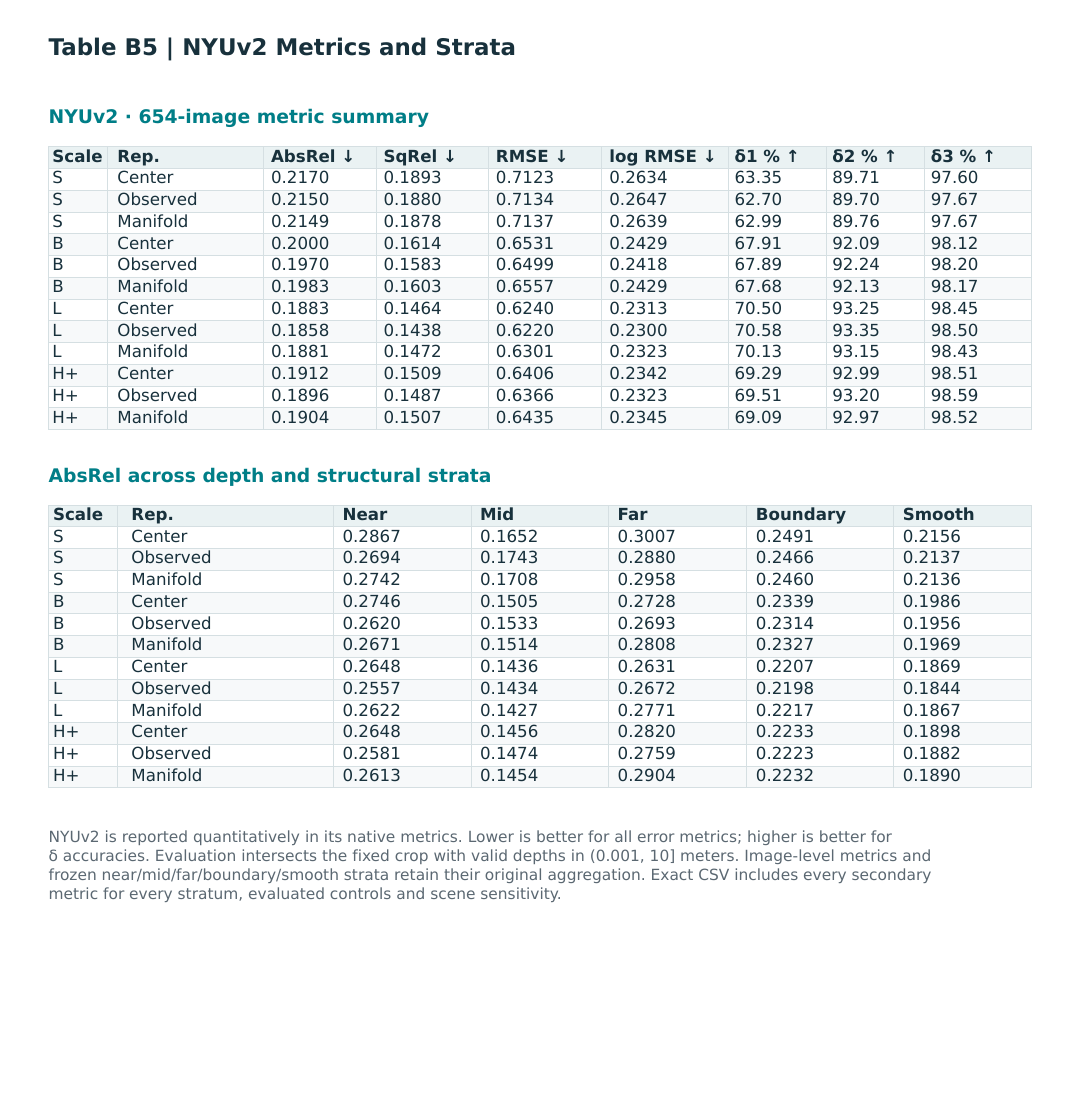}}
  \caption{\textbf{NYUv2 metrics and strata.} Lower is better for depth error metrics and higher is better for $\delta$ accuracy. All values retain their original image-level aggregation.}
  \label{tab:b5}
\end{table}

\begin{table}[H]
  \centering
  \PMTTcenterasset{\includegraphics[width=\PMTTAssetWidth]{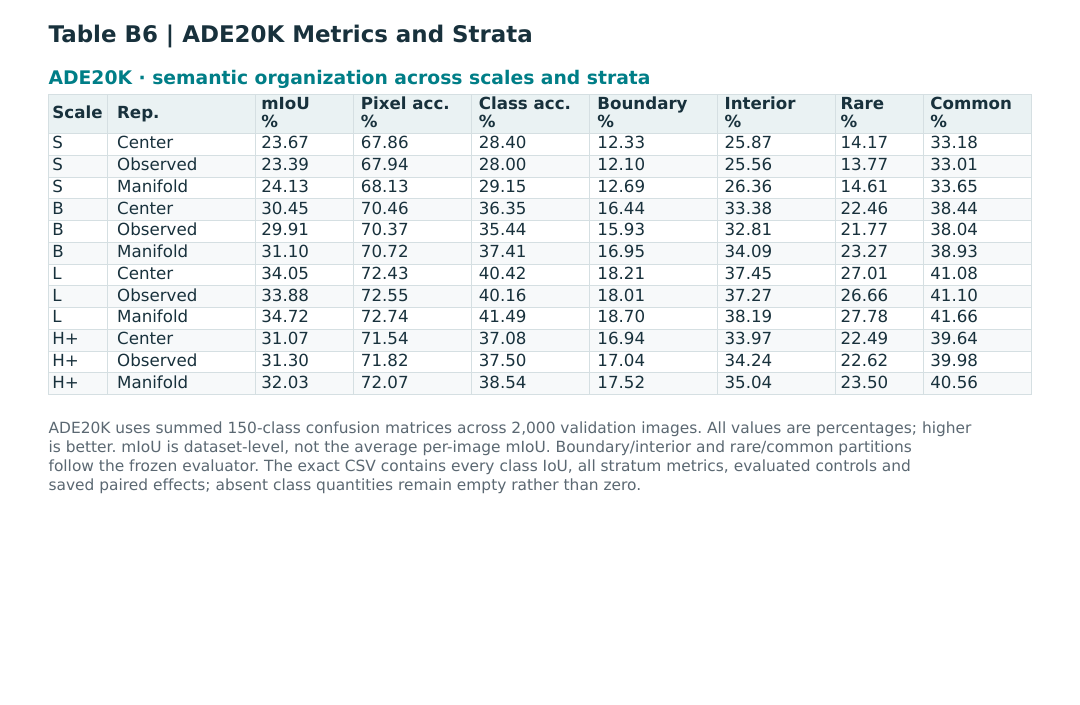}}
  \caption{\textbf{ADE20K metrics and strata.} Dataset-level mIoU, pixel accuracy, mean class accuracy, boundary/interior, and rare/common partitions are reported from summed 150-class confusion matrices.}
  \label{tab:b6}
\end{table}

\FloatBarrier
\section{Extended Pixel-Level Visualizations}
\label{app:qualitative}
\setcounter{figure}{0}
\renewcommand{\thefigure}{C\arabic{figure}}
The selected cases visualize strong positive mechanisms. They do not estimate prevalence. Population-level conclusions come from the complete quantitative matrix and paired statistics.

\begin{figure}[H]
  \centering
  \PMTTcenterasset{\includegraphics[page=1,width=\PMTTAssetWidth]{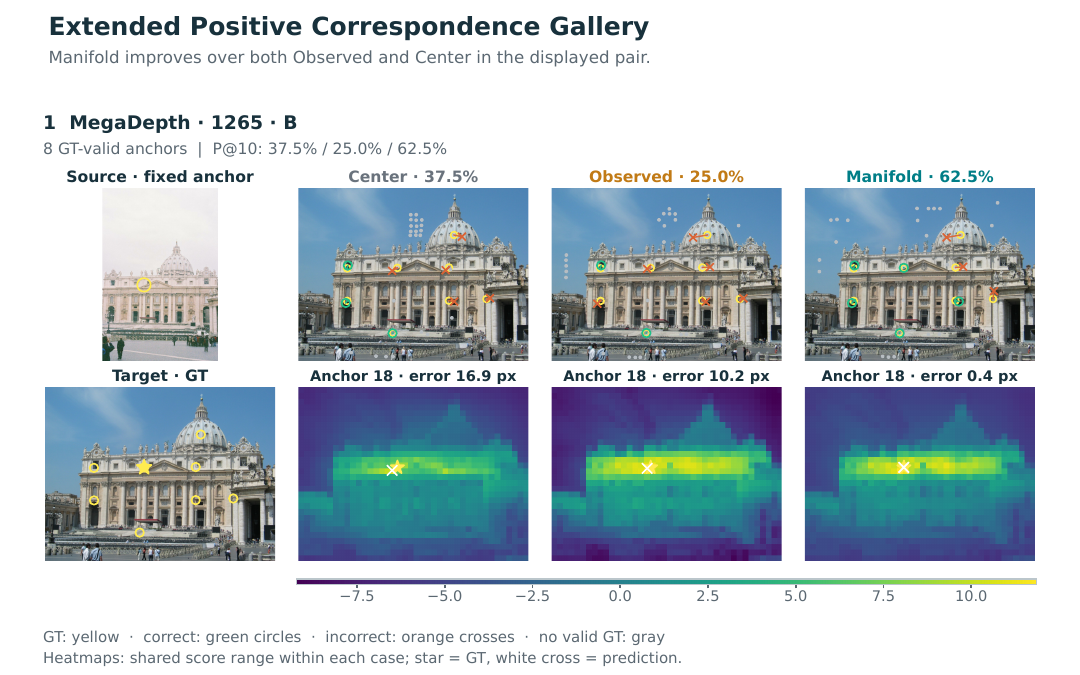}}
  \caption{\textbf{Extended positive correspondence gallery, page 1 of 4.} Additional cases extend the visible correspondence gains beyond \figref{fig:qualitative} without repeating its examples.}
  \label{fig:c1}
\end{figure}
\begin{figure}[H]\ContinuedFloat
  \centering
  \PMTTcenterasset{\includegraphics[page=2,width=\PMTTAssetWidth]{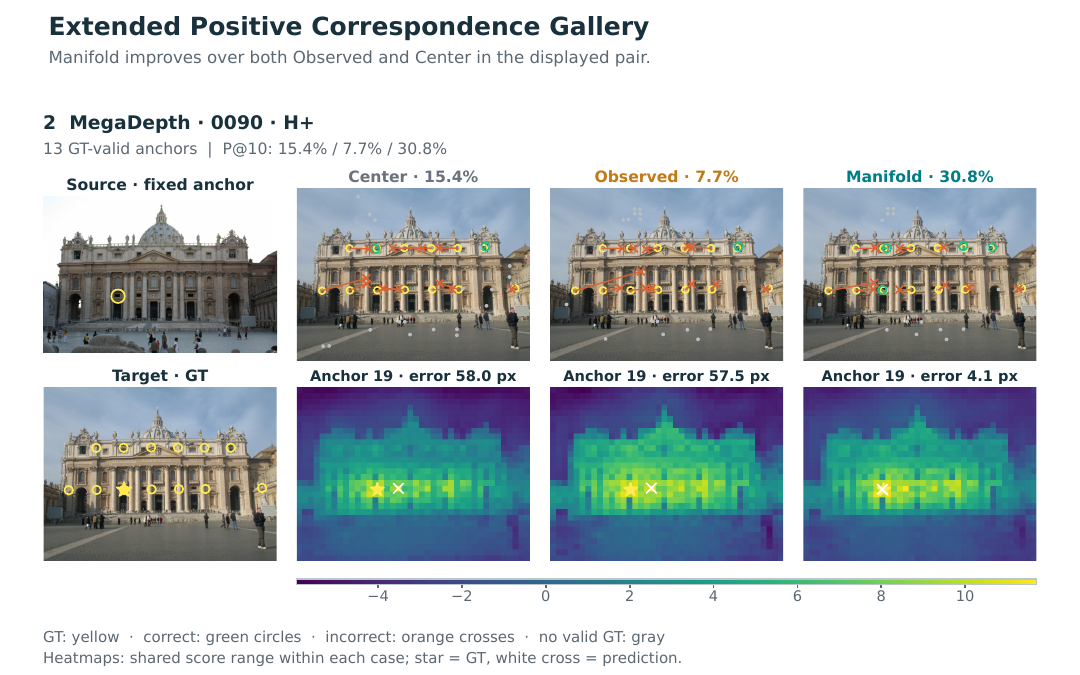}}
  \caption{\textbf{Extended positive correspondence gallery, page 2 of 4.}}
\end{figure}
\begin{figure}[H]\ContinuedFloat
  \centering
  \PMTTcenterasset{\includegraphics[page=3,width=\PMTTAssetWidth]{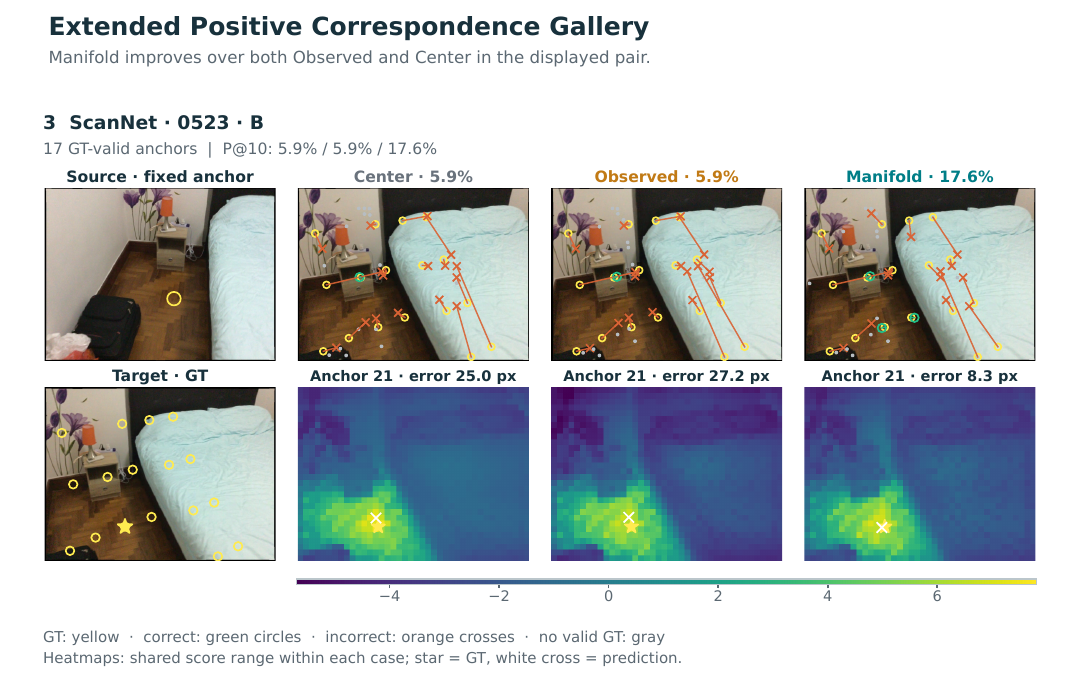}}
  \caption{\textbf{Extended positive correspondence gallery, page 3 of 4.}}
\end{figure}
\begin{figure}[H]\ContinuedFloat
  \centering
  \PMTTcenterasset{\includegraphics[page=4,width=\PMTTAssetWidth]{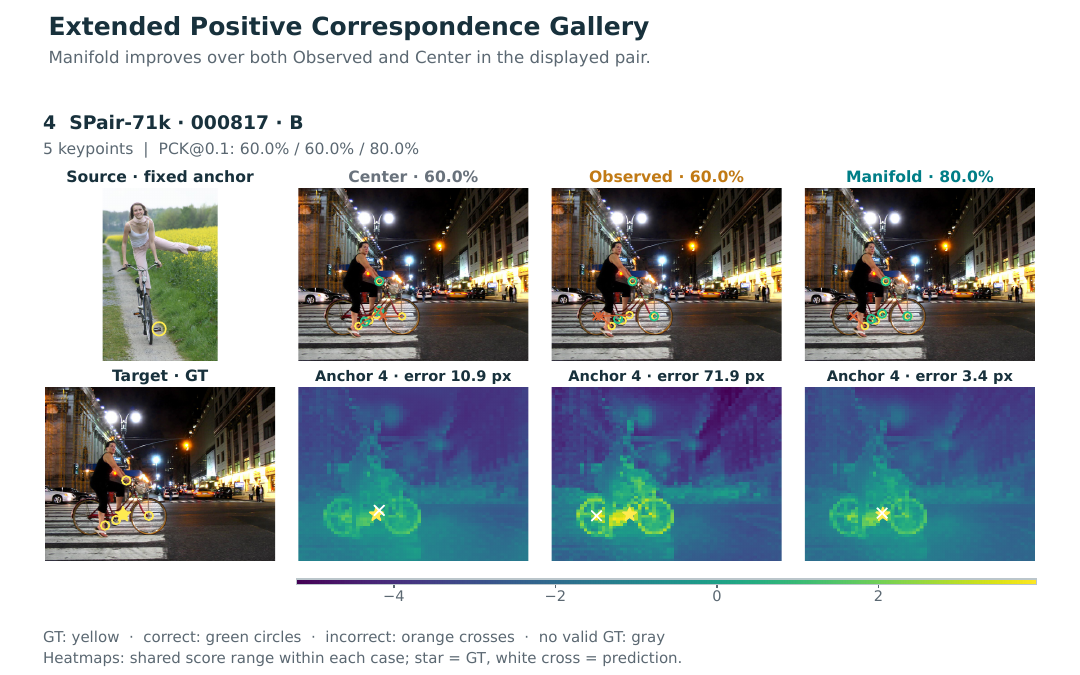}}
  \caption{\textbf{Extended positive correspondence gallery, page 4 of 4.}}
\end{figure}

\begin{figure}[H]
  \centering
  \PMTTcenterasset{\includegraphics[page=1,width=\PMTTAssetWidth]{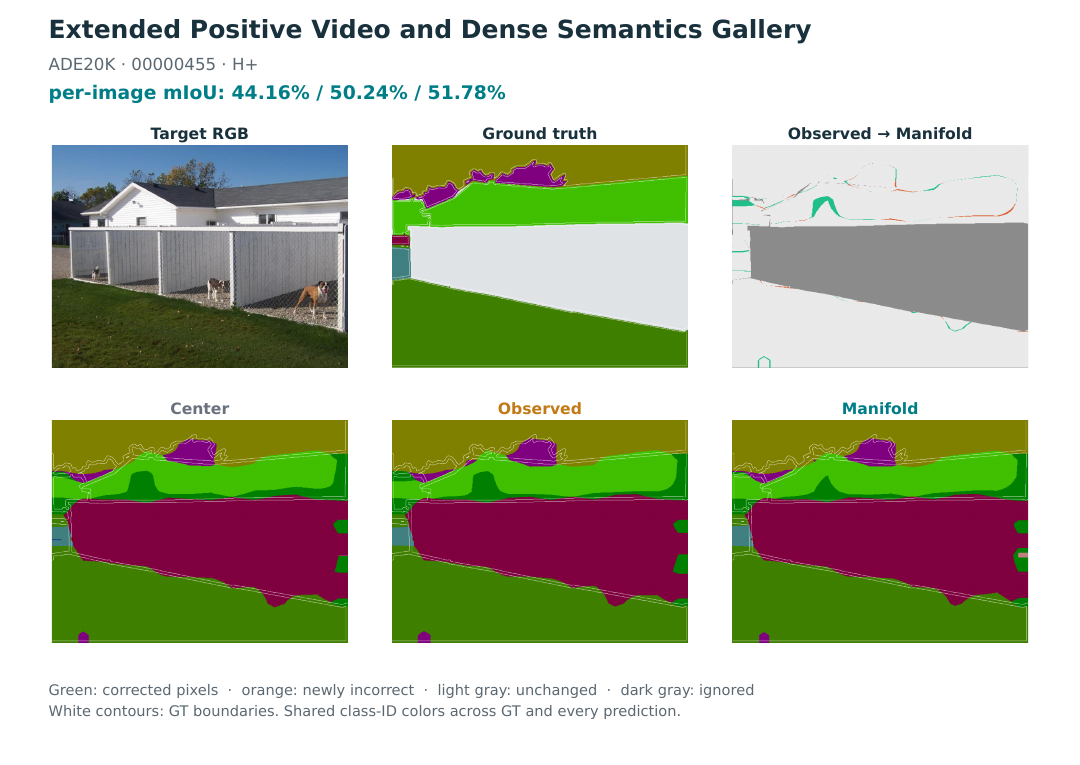}}
  \caption{\textbf{Extended positive video and dense-semantics gallery, page 1 of 5.} ADE20K and DAVIS examples show cases in which \Manifold improves over both \Observed and \Center. Green pixels become correct, orange pixels become incorrect, and gray pixels are unchanged or ignored.}
  \label{fig:c2}
\end{figure}
\begin{figure}[H]\ContinuedFloat
  \centering
  \PMTTcenterasset{\includegraphics[page=2,width=\PMTTAssetWidth]{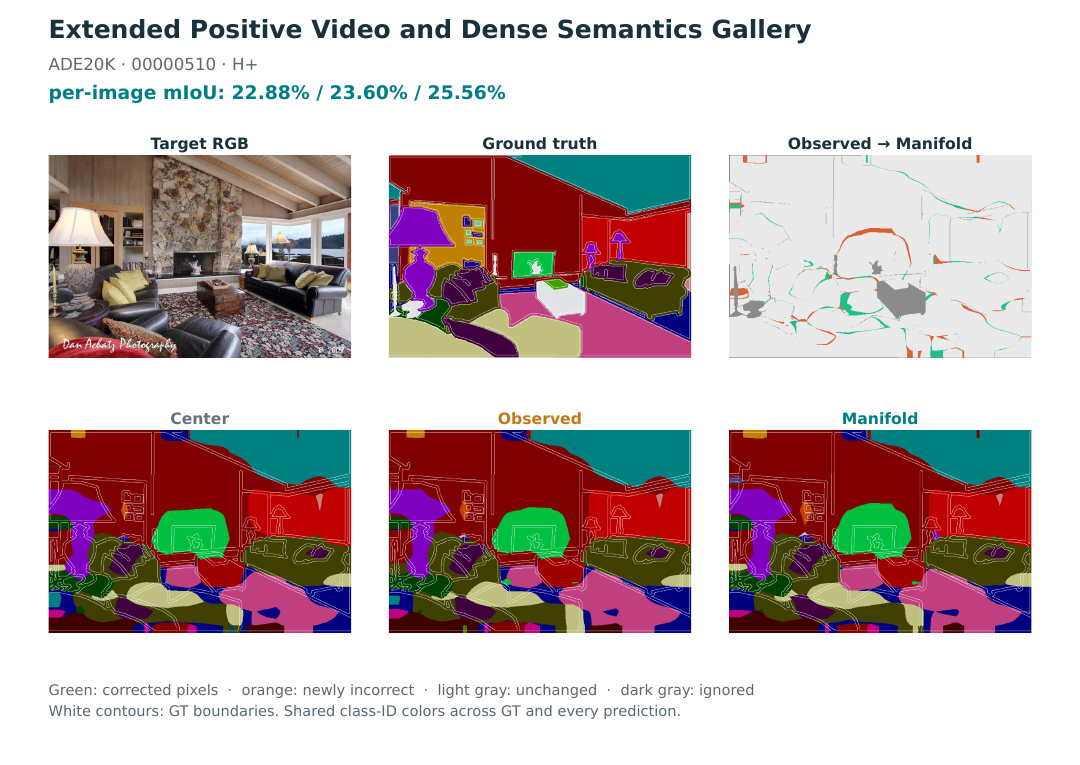}}
  \caption{\textbf{Extended positive video and dense-semantics gallery, page 2 of 5.}}
\end{figure}
\begin{figure}[H]\ContinuedFloat
  \centering
  \PMTTcenterasset{\includegraphics[page=3,width=\PMTTAssetWidth]{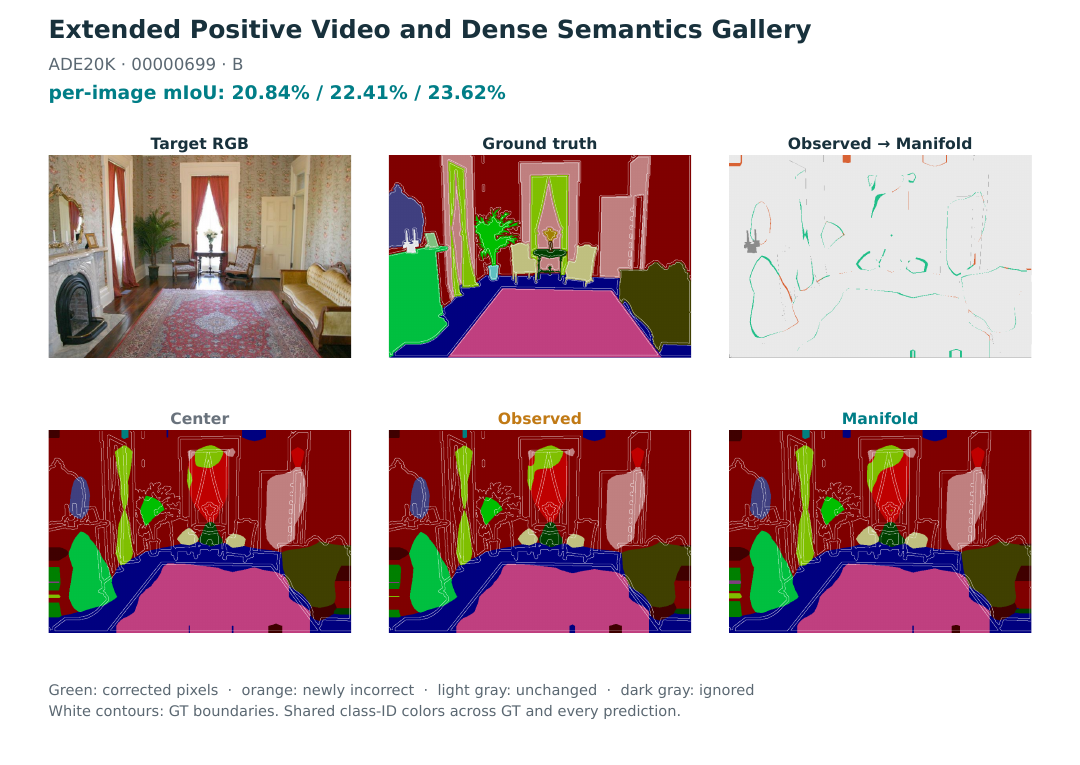}}
  \caption{\textbf{Extended positive video and dense-semantics gallery, page 3 of 5.}}
\end{figure}
\begin{figure}[H]\ContinuedFloat
  \centering
  \PMTTcenterasset{\includegraphics[page=4,width=\PMTTAssetWidth]{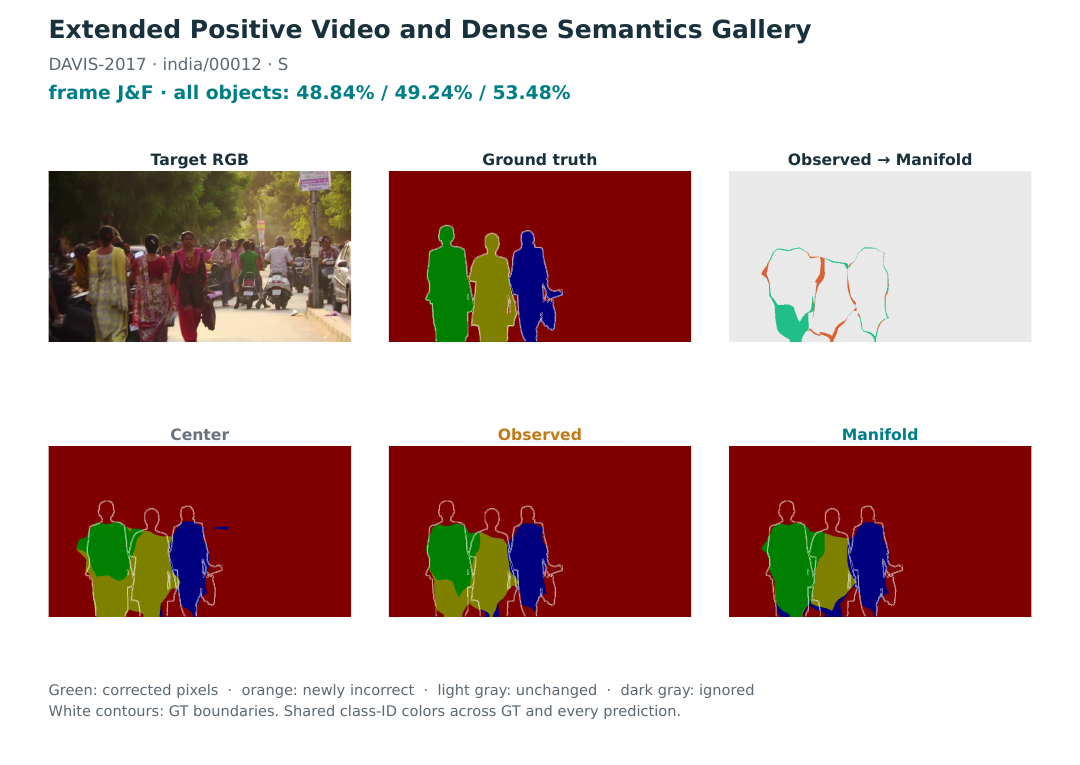}}
  \caption{\textbf{Extended positive video and dense-semantics gallery, page 4 of 5.}}
\end{figure}
\begin{figure}[H]\ContinuedFloat
  \centering
  \PMTTcenterasset{\includegraphics[page=5,width=\PMTTAssetWidth]{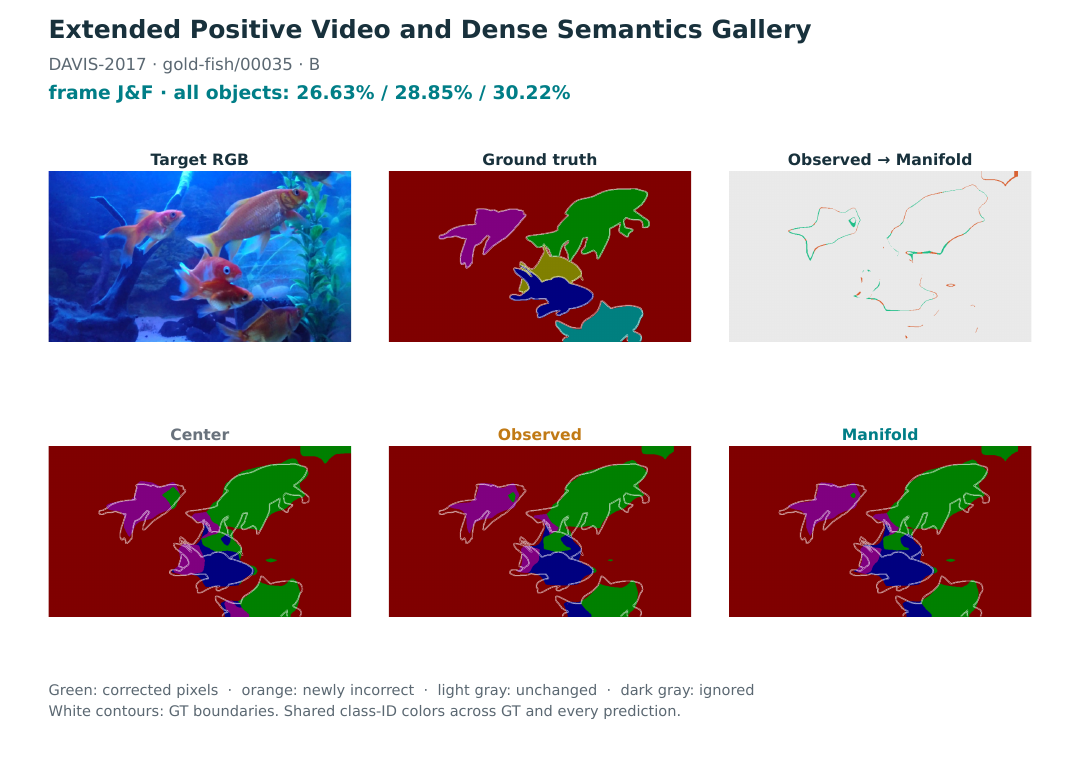}}
  \caption{\textbf{Extended positive video and dense-semantics gallery, page 5 of 5.}}
\end{figure}

\FloatBarrier
\section{Reproducibility and Open Research Scope}
\label{app:reproducibility}
\paragraph{Reproducible components.}
The representation can be reconstructed from a public DINOv3 backbone, the canonical patch identity, the deterministic Dense Orbital View parameters, the Visible/Hidden split, and the rank-8 bounded factor construction. Downstream reproduction additionally requires official datasets, fixed manifests, task-axis parameters or train-side fitting procedures, and the task-native evaluators described above.

\paragraph{Public release scope.}
A clean implementation and lightweight reproducibility assets can contain source code, configs, manifests, cohort identities, reader coefficients, dense-task axes, result CSVs, and table-reproduction inputs. Third-party datasets and DINOv3 weights should be obtained from their original sources. Full feature caches and private execution infrastructure are unnecessary for scientific reproduction.

\paragraph{Open directions.}
The current work gives a clean origin point rather than a fixed endpoint. Future research can learn acquisition policies, train patch-level Visible-to-Hidden objectives, replace the analytic factor with nonlinear or backbone-native organization, compress a state family into a compact code, and study observation-conditioned dynamics in video and 3D. The main scientific constraint is stable representation ownership: richer states must remain states of the same patch rather than an unstructured neighborhood of other patches.

\FloatBarrier

\end{document}